\documentclass[sigconf]{acmart}

\usepackage{algpseudocode}
\usepackage{mathtools}
\usepackage{balance}
\usepackage{subfigure}
\usepackage[flushleft]{threeparttable}
\usepackage{booktabs}
\usepackage{url}
\usepackage{graphicx}
\usepackage{multirow}
\usepackage{comment}
\usepackage[ruled,linesnumbered]{algorithm2e}

\usepackage{xcolor}

\AtBeginDocument{%
  \providecommand\BibTeX{{%
    \normalfont B\kern-0.5em{\scshape i\kern-0.25em b}\kern-0.8em\TeX}}}

\copyrightyear{2021}
\acmYear{2021}
\setcopyright{acmcopyright}\acmConference[SIGMOD '21]{Proceedings of the 2021 International Conference on Management of Data}{June 18--27, 2021}{Virtual Event, China}
\acmBooktitle{Proceedings of the 2021 International Conference on Management of Data (SIGMOD '21), June 18--27, 2021, Virtual Event, China}
\acmPrice{15.00}
\acmDOI{10.1145/3448016.3452779}
\acmISBN{978-1-4503-8343-1/21/06}

\begin{document}
\fancyhead{}

\title{RobustPeriod: Robust Time-Frequency Mining for Multiple Periodicity Detection}

\author{Qingsong Wen, Kai He, Liang Sun}
\affiliation{%
  \institution{DAMO Academy, Alibaba Group, Bellevue, USA}
%   \streetaddress{Anonymous address}
%   \city{Bellevue}
%   \state{WA, USA}
  \postcode{Anonymous postcode}}
\email{{qingsong.wen, kai.he, liang.sun}@alibaba-inc.com}

% \institution{Cloud-Computing Platform, Alibaba Group}
\author{Yingying Zhang, Min Ke, Huan Xu}
\affiliation{%
  \institution{Alibaba Group, Hangzhou, China}
%   \streetaddress{Anonymous address}
%   \city{Hangzhou}
%   \state{China}
  \postcode{Anonymous postcode}}
% \email{dawu@taobao.com,{congrong.zyy, huan.xu}@alibaba-inc.com}
\email{congrong.zyy@alibaba-inc.com, dawu@taobao.com, huan.xu@alibaba-inc.com}

\renewcommand{\shortauthors}{Qingsong Wen, Kai He, Liang Sun, et al.}


\begin{abstract}
% (Note: 8 pages for Data Science and Engineering papers in SIGMOD 2021, although we will allow an unlimited number of pages for the bibliography.)

Periodicity detection is a crucial step in time series tasks, including monitoring and forecasting of metrics in many areas, such as IoT applications and self-driving database management system. In many of these applications, multiple periodic components exist and are often interlaced with each other. Such dynamic and complicated periodic patterns make the accurate periodicity detection difficult. In addition, other components in the time series, such as trend, outliers and noises, also pose additional challenges for accurate periodicity detection. In this paper, we propose a robust and general framework for multiple periodicity detection. Our algorithm applies maximal overlap discrete wavelet transform to transform the time series into multiple temporal-frequency scales such that different periodic components can be isolated. We rank them by wavelet variance, and then at each scale detect single periodicity by our proposed Huber-periodogram and Huber-ACF robustly. We rigorously prove the theoretical properties of Huber-periodogram and justify the use of Fisher’s test on Huber-periodogram for periodicity detection. To further refine the detected periods, we compute unbiased autocorrelation function based on Wiener-Khinchin theorem from Huber-periodogram for improved robustness and efficiency. Experiments on synthetic and real-world datasets show that our algorithm outperforms other popular ones for both single and multiple periodicity detection.

\end{abstract}

\ccsdesc[500]{Mathematics of computing~Time series analysis}
\ccsdesc[500]{Information systems~Data mining}
\ccsdesc[500]{Information systems~Web mining}

\keywords{time series; periodicity detection; multiple periodicity; periodogram; ACF; database monitoring}



\maketitle

% \vspace{-0.5cm}
\section{Introduction}

Many time series are characterized by repeating cycles, or periodicity. For example, many human activities show periodic behavior, such as the cardiac cycle and the traffic congestion in daily peak hours. As periodicity is an important feature of time series, periodicity detection is crucial in many time series tasks, including time series similarity search~\cite{vlachos2004identifying,toyoda2013pattern}, forecasting~\cite{papadimitriou2003adaptive,yuan2020effective,faloutsos2019classical,iBTune19,Ma2018}, anomaly detection~\cite{jingkun20_TAD,rasheed2013framework}, decomposition~\cite{STL_cleveland1990stl,RobustSTL_wen2018robuststl,FastRobustSTL_wen2020,yang2021robuststl,dokumentov2020str}, classification~\cite{cPD_vlachos2005Autoperiod}, and compression~\cite{ts:multi-scale:compression}. Specifically, in forecasting tasks the prediction accuracy can be significantly improved by utilizing the periodic patterns~\cite{xu2021seasonforecast,lai2018modeling,higginson2020database}. Furthermore, periodicity detection plays an important role in resource auto-scaling. For example, the workloads of database and cloud computing often exhibit notable periodic patterns~\cite{higginson2020database,atikoglu2012workload,cortez2017resource,calzarossa2016workload,SPAR08}. By identifying periodic workloads, we can perform effective auto-scaling of resources in various scenarios, including virtual machine management in database management~\cite{taft2018p,jindal2019peregrine} and cloud computing\cite{mei2020turbine,jyothi2016morpheus}, leading to significantly less resource usage.

%Accurate periodicity detection not only enables better processing of the single time series, it also facilitates the study of multiple time series. For example, we can define similarity based on periodicity patterns between multiple time series, which can be used in further analysis such as classification and clustering. 
% forecasting~\cite{papadimitriou2003adaptive,xu2021seasonforecast,yuan2020effective,faloutsos2019classical,theodosiou2011forecasting,higginson2020database,iBTune19,Ma2018},
% forecasting~\cite{prema2015time,theodosiou2011forecasting}
% clustering~\cite{hyndman2006:characterizeTS,vlachos2004identifying}

% Paragraph 2. Challenges in periodicity detection.
Due to the diversity and complexity of periodic patterns arising in different real-world applications, accurate periodicity detection is challenging. Periodicity generally refers to the repeated pattern in time series. However, sometimes the periodic component can be dynamic and deviate from the normal behavior. An example is the sales amount of an online retailer exhibiting the daily periodicity, which can change dramatically when big promotion happens such as black Friday~\cite{taft2018p}. In addition, when multiple periodic components exist, they are generally interlaced with each other, which makes identifying all periodic components more challenging. For example, the traffic congestion time series typically exhibits daily and weekly periodicities, but the weekly pattern may change when long weekend happens. The interlaced multiple periodic components are also observed in database workload capacity planning~\cite{higginson2020database}.  Furthermore, other components can interfere the periodicity detection, including trend, noises, and outliers. 
%The commonly used autocorrelation can be biased significantly and leading to false periodicity detection results. 
In particular, many existing methods fail when outliers in the data last for some time.

% Paragraph 3. Related work
Periodicity detection has been widely researched in a variety of fields, including data management~\cite{vlachos2004identifying,cortez2017resource,Telescope2020ICDE}, data mining~\cite{cPD_vlachos2005Autoperiod,elfeky2005warp,drutsa2017periodicity,Toller2019}, signal processing~\cite{Wang2006,tenneti2015nested}, statistics~\cite{Abdullah:testing:periodicity}, astronomy~\cite{Graham2013,graham2013using,suveges2015comparative}, bioinformatics~\cite{wichert2004identifying,yang2011lspr}, etc. Among these periodicity detection algorithms, two fundamental methods are: 1) frequency domain methods identifying the underlying periodic patterns by transforming time series into the frequency domain; 2) time domain methods correlating the signal with itself via autocorrelation function (ACF). Specifically, the discrete Fourier transform (DFT) converts a time series from the time domain to the frequency domain, resulting in the so-called periodogram which encodes the strength at different frequencies. Usually the top-$k$ dominant frequencies are investigated to find the frequencies corresponding to periodicities. The periodogram is easy to threshold for dominant period but it suffers from the so-called spectral leakage~\cite{cPD_vlachos2005Autoperiod}, which causes frequencies not integer multiples of DFT bin width to disperse over the spectrum. Also the periodogram is not robust to abrupt trend changes and outliers. On the other hand, ACF can identify dominant period by finding the peak locations of ACF and averaging the time differences between them. Generally, ACF tends to reveal insights for large periods but is prone to outliers and noises. In particular, both DFT and ACF fail to process time series with multiple periodicities robustly and effectively. The periodogram may give misleading information when multiple interlaced periodicities exist. Note that multiples of the same period are also peaks in ACF, which leads to more peaks in the multiple periodicity setting. Thus, directly utilizing the properties of periodogram and ACF may lead to inaccurate periodicity detection results. Recently some algorithms combining DFT and ACF have been proposed~\cite{cPD_vlachos2005Autoperiod,puech2019fully,Toller2019}. Unfortunately, they cannot address all the aforementioned challenges.

%For the second method, i.e., the autocorrelation function, it computes the correlation between the time series and itself. 

% including astronomy~\cite{Graham2013,Otazu:wavelets}
% ~\cite{BLACKLEDGET200675,cPD_vlachos2005Autoperiod}
% our contributions
% state our methodology, also discribe how to overcome the above mentioned challenges. 
In this paper we propose a new periodicity detection method called  RobustPeriod %(time-frequent Period detection) 
to detect multiple periodicity robustly and accurately. To mitigate the side effects introduced by trend, spikes and dips, we introduce the Hodrick–Prescott (HP) trend filtering to detrend and smooth the data. To isolate different periodic components, we apply maximal overlap discrete wavelet transform (MODWT) to decouple time series into multiple levels of wavelet coefficients and then detect single periodicity at each level. To further speed up the computation, we propose a method to robustly calculate unbiased wavelet variance at each level and rank periodic possibilities. For those with highest possibility of periodic patterns, we propose a robust Huber-periodogram and apply Fisher's test to select the candidates of periodic lengths. Finally, we apply the  Huber-ACF to validate these period length candidates. By applying Wiener-Khinchin theorem, the unbiased Huber-ACF can be computed efficiently and accurately based on the Huber-periodogram, and then more accurate period length(s) can be detected. 

In summary, by applying MODWT and the unbiased wavelet variance, we can effectively handle multiple periodicities. The proposed Huber-periodogram and Huber-ACF can deal with impulse random errors with unknown heavy-tailed error distributions, leading to accurate periodicity detection results. We rigorously prove the theoretical properties of Huber-periodogram and justify the use of Fisher's test based on Huber-periodogram. Compared with various state-of-the-art periodicity detection algorithms, our RobustPeriod algorithm performs significantly better on both synthetic and real-world datasets.

\vspace{-0.3cm}
\section{Related Work}\label{sec:related}

%Periodicity detection is also referred as season length estimation~\cite{Toller2019}. 

% Traditional two types of periodicity detecteion algorithms

Most periodicity detection algorithms can be categorized into two groups: 1) frequency domain methods relying on periodogram after Fourier transform~\cite{vlachos2004identifying,drutsa2017periodicity,Telescope2020ICDE}; 2) time domain methods relying on ACF~\cite{radinsky2012modeling, Wang2006}. However, periodogram is not accurate when the period length is long or the time series is with sharp edges. Meanwhile, the estimation of ACF and the discovery of its maximum values can be affected by outliers and noises easily, leading to many false alarms in practice. 
% Note that periodicity detection is also referred as seasonality detection or season length estimation~\cite{Toller2019}, we use periodicity and seasonality in this paper equivalently.
% % periodogram-based approach
% In the frequency domain methods, the periodogram measures how periodic the time series is at different frequencies. If a time series does not contain a periodic component, then the peak values of periodogram are not significant. However, periodogram is not accurate when the period length is long or the time series is with sharp edges. In the time domain methods, the ACF measures the similarity between a sequence and itself for some lags. Intuitively, a time series is periodic with a period $T$ if it can be divided into equal segments with length $T$ sharing high similarities. In other words, the ACF whose lag is the multipliers of $T$ would be locally maximum or peaks. Based on this properties of ACF, numerous methods have been proposed, including the analysis of autocorrelation peaks~\cite{Wang2006}, autocorrelation zero-distance~\cite{Toller2019}. However, the estimation of ACF and the automated discovery of its maximum values can be affected by outliers and noises easily, leading to many false alarms in practice. 
% In addition, it is not easy to determine the threshold for autocorrelation in the time domain. 
%the power spectral density 
%In periodogram-based methods, or spectral density based methods.
%The periodogram-based methods can give a 
% ACF-based approaches
%  spectral leakage~\cite{BLACKLEDGET200675,cPD_vlachos2005Autoperiod}
% Combined approaches
Some methods have been proposed in the joint frequency-time domain to combine the advantages of both methods. In AUTOPERIOD~\cite{cPD_vlachos2005Autoperiod,Mitsa:2010:TDM:1809755}, it first selects a list of candidates in the frequency domain using periodogram, and then identifies the exact period in the time domain using ACF. The intuitive idea is that a valid period from the periodogram should lie on a hill of ACF. \cite{Toller2019} proposes an ensemble method called SAZED which combines multiple periodicity detection methods together. Compared with AUTOPERIOD, it selects the list of candidate periods using both frequency domain methods and time domain methods. Also different properties of autocorrelation of periodic time series are utilized to validate period. Unfortunately, it can only detect single period.

% Unfortunately, it is computationally expensive and can only detect single period. 

% Some other recent works on periodicity detection
Recently some other periodicity detection algorithms have been proposed in the field of data mining, signal processing and astrology. One improvement~\cite{cPD_parthasarathy2006robust} is proposed to handle non-stationary time series using a sliding window and track the candidate periods using a Kalman filter, but it is not universally applicable and not robust to outliers. 
In \cite{dPD_li2012mining}, a method immune to noisy and incomplete observations is proposed, but it can only handle binary sequences.  Recently, \cite{dPD_yuan2017detecting} proposes a method to detect multiple periodicities. Unfortunately, it only works on discrete event sequences.
% Some other types of periodicity detection algorithms in the literature include wavelet transform based method~\cite{Abdullah:testing:periodicity}, dynamic time warping based method~\cite{elfeky2005warp}, epoch folding~\cite{Graham2013}, etc.  

% Multiple periodicity detection
In multiple periodicity detection, a related topic is the pitch periodicity detection~\cite{com:Auditory:book} where multiple periodicities are associated with the fundamental frequency (F0). In fact, the periodic waveform repeats at F0 and can be decomposed into multiple components which have frequencies at multiples of the F0. In our scenarios, we may not have the fundamental frequency and the relationship between different frequencies can be more complicated. 

\begin{figure*}[!htb]
    \includegraphics[width=0.7\linewidth]{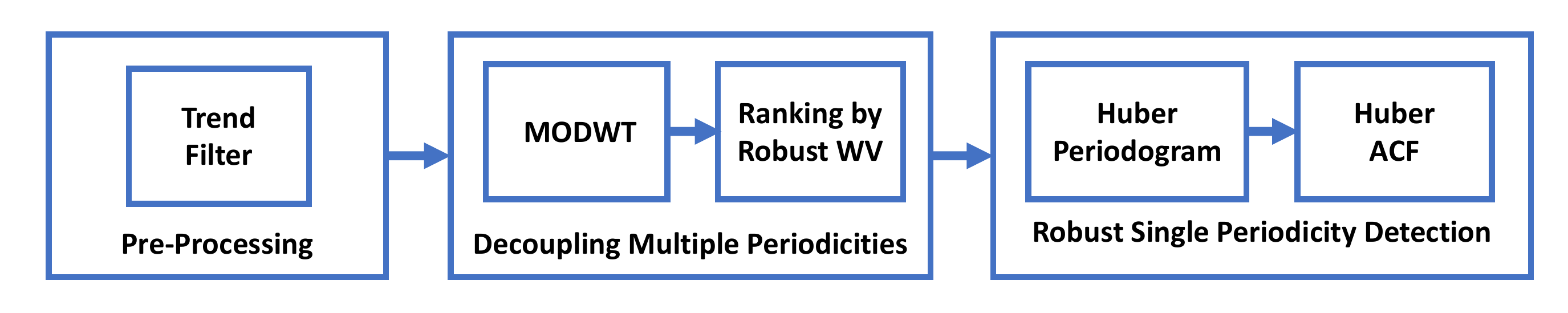}
    \vspace{-0.4cm}
    \caption{Framework of the proposed RobustPeriod algorithm.}
    \vspace{-0.4cm}
\label{fig:robustPeriod_Alg_frame}
\end{figure*}

\section{Methodology}

\subsection{Framework Overview}
We consider the following time series model with trend and multiple seasonality/periodicity as
\begin{equation}\label{eq:whole_model} 
y_t = \tau_t + \sum\nolimits_{i=1}^{m} s_{i,t} + r_t, \quad t = 0,1,\cdots,N-1
\end{equation}
where $y_t$ represents the observed time series at time $t$, $\tau_t$ denotes the trend component, $s_t= \sum_{i=1}^{m} s_{i,t}$ is the sum of multiple seasonal/periodic components with periods as $T_i, i=1,\cdots,m$, and $m$ is the number of periodic components. We use $r_t = a_t + n_t$ to denote the remainder part which contains the noise $n_t$ and possible outlier $a_t$. Our goal is to identify the number of the periodic components and each period length.

Intuitively, our periodicity detection algorithm first isolates different periodic components, and then verifies single periodicity by robust Huber-periodogram and the corresponding Huber-ACF. Specifically, RobustPeriod consists of three main components as shown in Fig.~\ref{fig:robustPeriod_Alg_frame}: 1) data preprocessing; 2) decoupling (potential) multiple periodicities by MODWT; 3) robust single periodicity detection by Huber-periodogram and Huber-ACF.

% RobustPeriod consists of five steps: 1) data preprocessing (normalization and detrend, and outlier removal); 2) robust Daubechies MODWT, which isolates different periodic components iteratively; 3) robust unbiased wavelet variance estimation to select the most promising level containing periodic patterns; 4) robust Huber-periodogram based Fisher's test for generating periodicity candidates; 5) robust ACF to refine and validate the final periodicities. In the following, we will delineate each step with details.

\subsection{Data Preprocessing}\label{sec:PeriodicityDetection}

%Here we first standardize the time series data to have a mean of 0 and a standard deviation of 1, which can remove the DC (direct current) component of the time series and facilitate the following DFT-based periodogram analysis. 

The complex time series in real-world may have varying scales and trends under the influence of noise and outliers. In the first step, we perform data preprocessing such as data normalization, detrending, and outlier processing. Here we highlight that the time series detrending is a key step 
as the trend component would bias the estimation of ACF, resulting in misleading periodic information. Specifically, we adopt Hodrick–Prescott (HP) filter~\cite{hodrick1997postwar} to estimate trend $\hat{\tau}_{t}$ due to its good performance and low computational cost:
%Then, HP filter is calculated on $\bar{y}_t$ to get the trend extracted signal $\hat{y}_t$, i.e.,
% \begin{align}\label{eq:hp_trend} 
% \hat{\tau}_{t} \! &=\! \underset{{\tau_{t}}}{\arg\min} \frac{1}{2}\sum_{t=0}^{N-1} ( \bar{y}_t - \tau_{t} )^2
% + \lambda\! \sum_{t=1}^{N-2} ( \tau_{t-1}\! - 2\tau_{t} + \tau_{t+1} )^2. 
% \end{align}
% After estimating the trend $\hat{\tau}_{t}$, we detrend the time series as
% $$
% \hat{y}_t \!&= \!y_t - \hat{\tau}_t.
% $$
\begin{align}\label{eq:hp_trend}  
\hat{\tau}_{t} \! =\! \underset{{\tau_{t}}}{\arg\min} \frac{1}{2}\!\sum\nolimits_{t=0}^{N-1} \!( {y}_t \!-\! \tau_{t} )^2
+ \lambda\! \sum\nolimits_{t=1}^{N-2}\! ( \tau_{t-1}\! \!-\! 2\tau_{t} \!+\! \tau_{t+1} )^2. 
\end{align}
After estimating the trend $\hat{\tau}_{t}$, the detrended time series $\hat{y}_t = y_t - \hat{\tau}_t$ is further processed to coarsely remove extreme outliers by ${y}'_t \!= \Psi(\frac{\hat{y}_t-\mu}{s})$ as in~\cite{Durre2015}, where $\mu$ and $s$ are the median and mean absolute deviation (MAD) of $\hat{y}_t$, respectively, and $\Psi(x)=sign(x)min(|x|,c)$ with tuning parameter $c$.

\subsection{Robust MODWT: Decouple Multiple Periodicities}
\subsubsection{Daubechies MODWT for time series decomposition}
% Discrete wavelet transform (DWT) maps a signal to another representation by recursively applying two discrete wavelet functions, namely the scaling function (called father wavelet) that reconstructs the lower frequency (smooth) part, and the wavelet function (called the mother wavelet) that describes the higher frequency part. The commonly used discrete wavelet functions include Harr wavelets and Daubechies wavelets~\cite{Daubechies1992}.

% Discrete wavelet transform (DWT) can decompose time series into multiple time series at different scales and frequencies to facilitate time series analysis. However, the traditional DWT suffers from two drawbacks, including the requirement of sample size being a power of two and the fact that the wavelet and scaling coefficients are not shift invariant. This motivates the use of a variant of DWT, i.e., the maximal overlap DWT (MODWT), due to the following reasons: 1) ability to handle any sample size; 2) increased resolution at coarser scales; 3) a more asymptotically efficient wavelet variance estimator than DWT.

We adopt maximal overlap discrete wavelet transform (MODWT) to decompose the input time series into multiple time series at different levels to facilitate periodicity detection. The motivation to use MODWT instead of DWT is due to the following advantages of MODWT: 1) ability to handle any sample size; 2) increased resolution at coarser scales; 3) a more asymptotically efficient wavelet variance estimator than DWT; 4) can handle non-stationary time series and non-Gaussian noises more effectively.

%To mitigate the extreme values in time series in wavelet transform, we down-weight those values with the following transformation \cite{Bustos86_robustARMA} for time series $\hat{y}_t$ 
%$$
%y'_t= \psi_{c} \left( \frac{\hat{y}_t - \text{Med}(\hat{y}_t)}{1.4826 \cdot\text{MAD}(\hat{y}_t)}\right),
%$$
%where $\text{Med}(x) $ is the sample median, $\text{MAD}(x) $ is the median absolute deviation defined as $\text{MAD}(x) = \text{Med}(|x-\text{Med}(x)|) $, $\psi_{c}(x) = \text{sgn}(x)\text{min}(|x|,c)$ is the Huber transformation function and the tuning parameter is commonly set $c=1.37$ \cite{Durre2015}. The constant $1.4826$ is used to adjust the MAD as a consistent estimator for the estimation of the standard deviation.

\begin{figure}[!t]
    \includegraphics[width=0.9\linewidth]{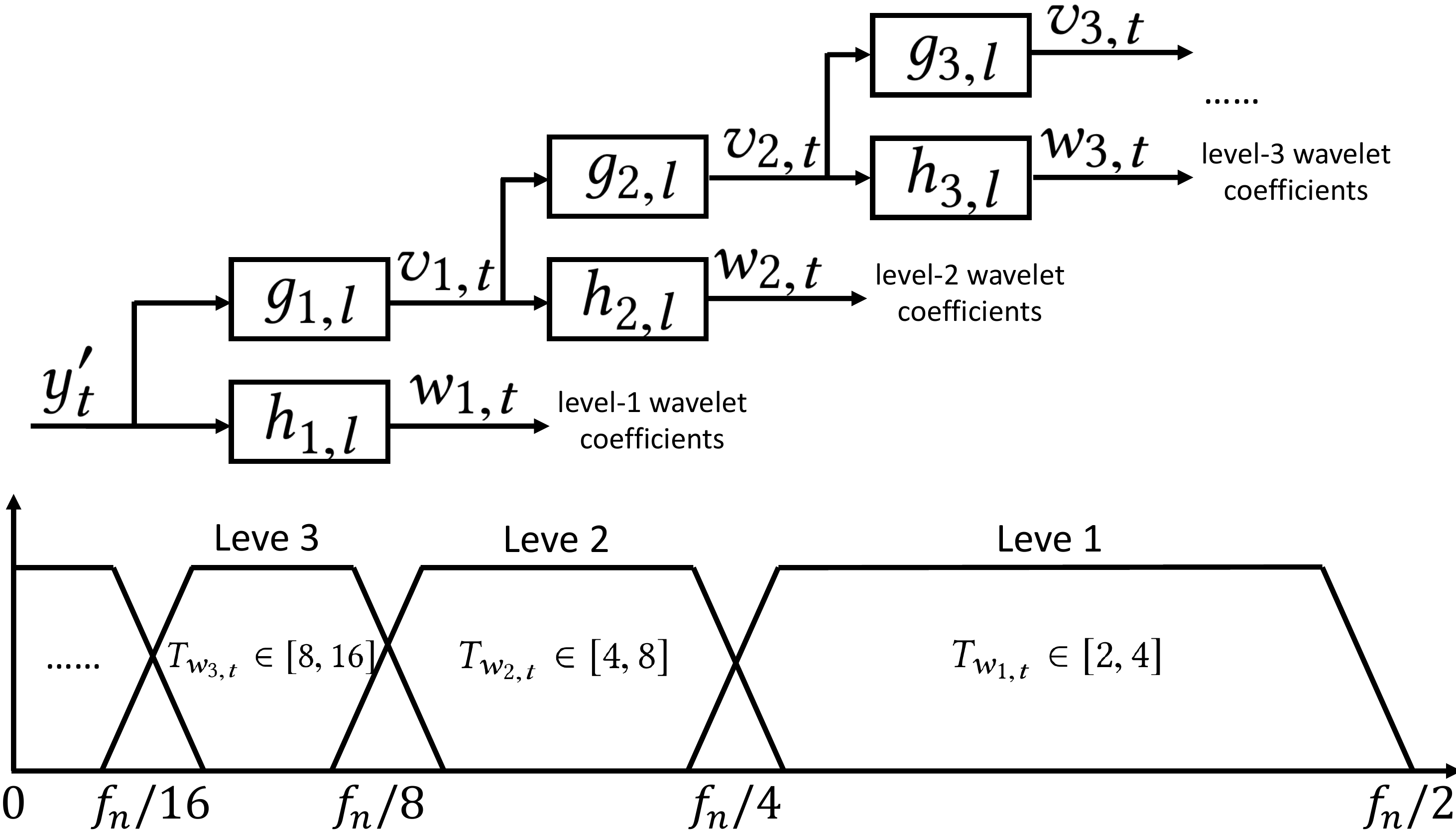}
    \vspace{-0.2cm}
    \caption{MODWT for decoupling multiple periodicities.}
    \vspace{-0.5cm}
\label{fig:modwt_demo}
\end{figure}

Here we adopt the common Daubechies based MODWT~\cite{percival2000wavelet,Daubechies1992} for time series analysis. When MODWT is performed on time series $y'_t$, the $j$th level wavelet and scaling coefficients ${w}_{j,t}$ and ${v}_{j,t}$ are
\begin{equation}\label{eq:modwt-eq} 
w_{j,t} \!= \!\!\sum\nolimits_{l=0}^{L_j-1}h_{j,l}y'_{t-l~mod~N}, \quad
v_{j,t} \!=\!\! \sum\nolimits_{l=0}^{L_j-1}g_{j,l}y'_{t-l~mod~N},  %t=0,1,\cdots,N-1
\end{equation}
where $\{h_{j,l}\}_{l=0}^{L_j-1}$, $\{g_{j,l}\}_{l=0}^{L_j-1}$ are $j$th level wavelet filter and scaling filter, respectively, and the filter width is $L_j = (2^j-1)(L_1-1)+1$ with $L_1$ as the width of unit-level Daubechies wavelet coefficients~\cite{Daubechies1992}.
Note that the wavelet filter $h_{j,l}$ in Eq.~\eqref{eq:modwt-eq} performs band-pass filter with nominal passband as $1/2^{j+1} \leq |f| \leq 1/2^j$. Therefore, if there is a periodic component of the time series $y'_t$ located in the nominal passband $1/2^{j+1} \leq |f| \leq 1/2^j$, this periodic component would be filtered into the $j$th level wavelet coefficient. Therefore, we can decouple multiple periodicities by adopting MODWT where the possible period length of $j$th level wavelet coefficients is within length of $[2^j, 2^{j+1}]$, as illustrated in Fig.~\ref{fig:modwt_demo}.

In real-world scenarios with outliers and noise, the time series is usually a non-Gaussian process with some degree of memory and correlation. But the MODWT can overcome these shortcomings to some extent, since the wavelet coefficients from MODWT are approximately Gaussian~\cite{Mallows1967}, uncorrelated and stationary~\cite{Zhu2014}. These properties would improve the performance of periodicity detection.

\subsubsection{Robust Unbiased Wavelet Variance}
Besides decoupling multiple periods of time series, another benefit of MODWT is that the corresponding wavelet variance estimation helps to locate the periodic component in the frequency bands as it is actually a rough estimate of the PSD. Thus, we can rank possible single periodic components by their corresponding wavelet variances.

%the corresponding wavelet variance provides hints of the periodic structure in time series, which can offer an intuitive explanation of how periodic components are located in time series. 

For level $J_0$ decomposition, based on the energy preserving of MODWT, we have
$
||\mathbf{y}'||^2 = \sum_{j=1}^{J_0}||\mathbf{w}_j||^2 + ||\mathbf{v}_{J_0}||^2, 
$
which leads to wavelet variance decomposition as
$
\hat{\sigma}^2_{\mathbf{y}'} = \sum_{j=1}^{J_0} \hat{\sigma}^2_{\mathbf{w}_j} + \hat{\sigma}^2_{\mathbf{v}_{J_0}}
$
where $\hat{\sigma}^2_{\mathbf{w}_j}$, $\hat{\sigma}^2_{\mathbf{v}_{J_0}}$ are the $j$th level empirical wavelet variance and level $J_0$ empirical scaling variance, respectively.
If $y'_t$ is stationary, then 
$
\hat{\sigma}^2_{\mathbf{y}'} = \sum_{j=1}^{\infty} \hat{\sigma}^2_{\mathbf{w}_j}.
$
Therefore, wavelet variance provides a scale-based analysis of variance  for time series, which can offer an intuitive explanation of how a time series is structured. 
% The wavelet variance provides a scale-based analysis of variance (ANOVA) for time series, which can offer an intuitively sensible explanation of how a time series is structured. 

We adopt biweight midvariance as the estimation of wavelet variance due to its robustness and efficiency~\cite{wilcox2011introduction}. Furthermore, the first $L_j-1$ wavelet coefficients are excluded for the aim of unbiased variance estimation, since the wavelet transform introduces periodic extension as defined in Eq.~\eqref{eq:modwt-eq}. Therefore, we use the following formulation for robust unbiased estimation of wavelet variance:
\begin{equation}
\hat{\nu}^2_{\mathbf{w}_j} = \frac{M_j \sum\nolimits_{t=L_j-1}^{N-1} (w_{j,t} - \text{Med}(w_{j,t}))^2 (1-u_t^2)^4 I(|u_t|<1) }
{\left(\sum\nolimits_{t=L_j-1}^{N-1} (1-u_t^2)(1-5u_t^2) I(|u_t|<1) \right)^2},
\end{equation}
where $I(x)$ is the indicator function, $M_j=N-L_j+1$ is the number of nonboundary coefficients at the $j$th level,
$\text{Med}(w_{j,t})$ and $\text{MAD}(w_{j,t})$ are median and mean absolute deviation (MAD) of $w_{j,t}$, respectively, and $% \mathcal{M}(w_{j,t})
u_t = (w_{j,t}-\text{Med}(w_{j,t}))/(9 \cdot \text{MAD}(w_{j,t}))
$.

%The wavelet variance estimation helps to locate the periodic component of time series in the frequency bands, since it is actually a rough estimate of the PSD. 

For the $j$th level wavelet coefficients, its wavelet variance is approximately equal to the integral of PSD at corresponding nominal octave passband, i.e., 
$
\hat{\nu}^2_{\mathbf{w}_j} \approx  \int_{1/2^{j+1} \leq |f| \leq 1/2^j}^{} S_{\mathbf{y}'}(f) df.
$
It can be concluded that if there is a periodic component filtered into the $j$th level wavelet coefficient, a large value of $\hat{\sigma}^2_{\mathbf{w}_j}$ would be expected. 
Therefore, we only use the levels of wavelet coefficients occupying the dominating energy based on wavelet variance for single periodicity detection to speed up the computation.
Furthermore, we rank the wavelet coefficients based on the wavelet variances. 
Then, the order of single-period detection in each wavelet coefficient can follow this ranking to output the most significant periods first.

\subsection{Robust Single Periodicity Detection}\label{sec:RobustFisher}

\subsubsection{Robust Huber-Periodogram based Fisher's Test for Generating Periodicity Candidates}\label{sec:RobustFisher}
%\subsection{Robust Fisher's Test with Huber-Periodogram} \label{sec:RobustMPeriodogram}

In this subsection we design a robust Huber-periodogram based Fisher's test for improved single periodicity detection. We also provide the theoretical properties of the Huber-periodogram suitable for Fisher's test.

% In this subsection we discuss how to generate the period candidate list by computing the robust Huber-periodogram and Fisher's test. We also investigate the theoretical property of Fisher's test after the Huber-periodogram.

%(intro) For each level of wavelet coefficient, we aim to detect the possible hidden single periodicity. 

First, we double the length of wavelet coefficient $\mathbf{w}_j$ of each level by padding $N$ zeros denoted as $\mathbf{x}_j=[\mathbf{w}^T_j, 0, \cdots, 0]^T$, where the length of $\mathbf{x}_j$ is $N'=2N$.
% $%\begin{equation}
% \mathbf{x}_j=[\mathbf{w}^T_j, \underbrace{0, \cdots, 0}_{N~\text{zeros}}]^T
% $%\end{equation}
The purpose of this padding operation is to obtain robust ACF through Huber-periodogram (will be shown later). 
%Besides, it also reduces the complexity of of ACF from $O(N^2)$ to $O(N\log N)$, and increases the accuracy of periodogram as well. 
In the following, we drop the level index $j$ for simplification.
To detect the dominant periodicity, Fisher's test \cite{fisher1929tests} defines the $g$-statistic as
% \begin{equation}\label{eq:Fisher_g-statistics} 
$g = {max_{k}{P_k}}/({\sum_{j=1}^{N}P_j}), k=1,2, \cdots, N$,
% \end{equation}
where $P_k$ is the periodogram based on DFT and it is defined as
\begin{equation}\label{eq:Periodogram} 
P_k \!=\! \left|\left| \text{DFT}\{x_t\}  \right|\right|^2 \!=\! \frac{1}{{N'}} \sum\nolimits_{t=0}^{N'-1} \! \left|\left| x_te^{-i2 \pi kt/N'}\right|\right|^2, \quad i=\sqrt{-1}.
\end{equation}
%If $x_t$ is the white Gaussian noise with variance $\sigma^2$, then the periodogram $P_1, \cdots, P_q$ are independent and follow the chi-square distribution: 
%$$
%P_k \sim (1/2)\sigma^2 \chi^2(2),  \quad k=1,\cdots,q. 
%$$
% where $i=\sqrt{-1}$. The Fisher's test in Eq.~\eqref{eq:Fisher_g-statistics} sets a null hypothesis $H_0$ that the time series is generated by a Gaussian white noise sequence, against the alternative hypothesis $H_1$ that the data is generated by a Gaussian white noise sequence with a superimposed deterministic periodic component. 
In Fisher's test, under the null hypothesis that the time series is Gaussian white noise with variance $\sigma^2$, the distribution of $P_k$ is a chi-square distribution with $2$ degrees of freedom, i.e., $P_k \sim (1/2)\sigma^2 \chi^2(2)$. Therefore, the distribution of $g$-statistic \cite{fisher1929tests} under null hypothesis is 
% \cite{fisher1929tests,Brockwell_TSbook_1991}
% \begin{equation}
$P(g \geq g_0) =1-\sum_{k=1}^{\lfloor 1/g_0 \rfloor } \frac{(-1)^{k-1}N!}{k!(N-k)!}(1-kg_0)^{N-1}$,
% \end{equation}
which gives a $p$-value to determine if a time series is periodic.
If this value is less than the predefined threshold $\alpha$, we reject the null hypothesis and conclude the time series is periodic with dominant period length as $N'/k$ where $k = \arg \max_k P_k$.

The Fisher's test with the original periodogram in Eq.~\eqref{eq:Periodogram} is vulnerable to outliers, so we adopt the robust M-periodogram~\cite{Katkovnik1998}  
\begin{equation}\label{eq:Periodogram_final} 
P_k^M = \frac{N'}{4} \left|\boldsymbol{\hat{\beta}}_{M}(k)\right|^2
= \frac{N'}{4} \left|   \underset{\boldsymbol{\beta} \in \boldsymbol{R}^2}{\arg\min}
~\!\gamma( \boldsymbol{\phi} \boldsymbol{\beta} -  \mathbf{x} ) \right|^2,
\end{equation}
where $\gamma(\mathbf{x}) = \sum_{t=0}^{N'-1} \gamma(x_t)$ is a robustifying loss function, $\mathbf{x}  = [x_0,x_1, \cdots, x_{N'-1}]^T$, and $\boldsymbol{\phi} = [\boldsymbol{\phi}_0,\boldsymbol{\phi}_1, \cdots, \boldsymbol{\phi}_{N'-1}]^T$ with harmonic regressor $\boldsymbol{\phi}_t  = [\cos(2 \pi kt/N'),\sin(2 \pi kt/N')]$. The M-periodogram with sum-of-squares loss $\gamma (x_t)=x_t^2$ is equivalent to the original periodogram in Eq.~\eqref{eq:Periodogram}, while the M-periodogram with least absolute deviation (LAD) loss $\gamma (x_t)=|x_t|$ is the LAD-periodogram~\cite{Katkovnik1998,Li2008}. In this paper, we instead adopt Huber loss~\cite{huber2011robust} in Eq.~\eqref{eq:Periodogram_final} to obtain Huber-periodogram. One reason is that Huber loss is a combination of sum-of-squares loss and LAD loss, which is not only robust to outliers bust also adaptive on different types of data~\cite{huber2011robust,WenRobustTrend19}. Another reasaon is that the Huber-periodogram can bring more robust ACF (will be shown later), which is beneficial for validating final periodicity. 
Specifically, the $\gamma(x_t)$ for Huber-periodogram is
\begin{equation}\label{eq:huber_loss} 
\gamma (x_t) = \gamma_{\zeta}^{hub} (x_t) =
\begin{cases} 
\frac{1}{2} x_t^2,       & |x_t| \leq \zeta\\
\zeta|x_t|- \frac{1}{2} \zeta^2, & |x_t| > \zeta
\end{cases}
\end{equation}
The Huber-periodogram in Eq.~\eqref{eq:Periodogram_final} can be efficiently solved by ADMM~\cite{boyd2011distributed} method, and the distribution of the $P_k^M$ has the following proposition:
\begin{proposition} \label{prop:HuberPeriodogram_dist}
Under some practical mild conditions for the time series $\mathbf{x}$, as $n \rightarrow \infty$, we have
% \begin{align}
% \sqrt{N} \text{vec} \{ \boldsymbol{\hat{\beta}}_{M}(k) \}_{k=1}^q &\stackrel{A}{\sim} N(\mathbf{0},2m_2\mathbf{S}) \\
% \{ P_k^M\}_{k=1}^N \stackrel{A}{\sim} \{(1/2)m_2S_p(k)\chi_2^2\}_{k=1}^N,
% P_k^M &\stackrel{A}{\sim} (1/2)m_2S_p(k)\chi_2^2,
% \end{align}
% where $\mathbf{S} := \text{diag}\{S_2(1),S_2(2),\dots,,S_2(q)\}$, the sign $\stackrel{A}{\sim}$ represents ``asymptotically distributed as".
% \end{proposition}
% \begin{align}
% P_k^M &\stackrel{A}{\sim} (1/2)m_2S_2\chi_2^2
% \end{align}
$P_k^M \stackrel{A}{\sim} (1/2)m_2S_2\chi_2^2$,
where the sign $\stackrel{A}{\sim}$ represents ``asymptotically distributed as", $m_2$ is the second moment of $\{\mathbf{x}_t\}$, and $S_2 := \sum_{\tau=-\infty}^{\infty} r_2(\tau) cos(\frac{2\pi k\tau}{n}) > 0$ with absolutely summable ACF $r_2(\tau)$ for process $\{g_2(\mathbf{x}_t)\}$ with $g_2(x) := |x|sgn(x)$.
\end{proposition} 

% \subsubsection{Proof of Proposition 3}
% \addtocounter{proposition}{2} % this is for appendix with right numbering
% \begin{proposition} \label{prop:HuberPeriodogram_dist}
\begin{proof}
Let $\boldsymbol{\hat{\beta}_M(k)}$ be the minimizer of the problem:
\begin{equation}\label{eq:huber-opt}
    \boldsymbol{\hat{\beta}_M(k)} := \text{arg min}_{\boldsymbol{\beta} \in \boldsymbol{R}^2} \sum\nolimits_{t=1}^{N'} \gamma( \boldsymbol{\phi}_t \boldsymbol{\beta} -  x_t ),
\end{equation}
where $\gamma(\cdot)$ is defined in \eqref{eq:huber_loss}. Assume that $\{x_t\}$ satisfies:
% the following conditions:
\begin{itemize}
    % \item[a)] The $\{x_t\}$ have a common univariate probability density function $f(x)$ which is bounded and satisfies $m_2 := E(|x_t|^2) = \int |x|^2 f(x) dx < \infty$.
    \item[a)] The $\{x_t\}$ have a probability density function $f(x)$ which is bounded and satisfies $m_2 := E(|x_t|^2) = \int |x|^2 f(x) dx < \infty$.
    \item[b)] The $\{x_t\}$ are $\phi$-mixing with mixing coefficients $\phi(\tau)$ satisfying $\sum_{\tau=1}^\infty \sqrt{\phi(\tau)} < \infty$.
    \item[c)] The process $\{g_2(x_t)~|~g_2(x) := |x|sgn(x)\}$ is stationary in 2nd moments with zero mean and absolutely summable ACF $r_2(\tau)$ such that
    $S_2 := \sum_{\tau=-\infty}^{\infty} r_2(\tau) cos(\frac{2\pi k\tau}{n}) > 0$.
    % \item[d)] For any fixed integer $q \ge 1$, let $\{\omega_1,\omega_2,\dots,\omega_q\} \subset (0,\pi)$ be a set of distinct values which may depend on $n$ but satisfies the condition such that
    % $n^{-1}\sum_{t=1}^n\mathbf{c}_t(\omega_j)\mathbf{c}_t^T(\omega_k) = (1/2)\delta_{j-k}\mathbf{I} + o(1)$ as $n \rightarrow \infty$, where $\{\delta_j\}$ is the Kronecker delta sequence such that $\delta_0 = 1$ and $\delta_j = 0$ for $j \neq 0$.
    \item[d)] Let $n_1$ be the length of subsequence $\{x_{t_1}\} \subset \{x_t\}, |x_{t_1}-\boldsymbol{\phi}_{t_1} \boldsymbol{\beta}|\le \zeta$ and $n_2$ be the length of subsequence $\{x_{t_2}\} \subset \{x_t\}, |x_{t_2}-\boldsymbol{\phi}_{t_2} \boldsymbol{\beta}|> \zeta$, we have $n_1 \ge n_2^2$.
\end{itemize}{}
Then as $N' \rightarrow \infty$, $\sqrt{N'} \boldsymbol{\hat{\beta}_M(k)} \stackrel{A}{\sim} N(\mathbf{0},2m_2\mathbf{S})$, and $P_k^M \stackrel{A}{\sim} (1/2)m_2S_2\chi_2^2$, where $P_k^M := \frac{N'}{4}||\boldsymbol{\hat{\beta}_M(k)}||^2$, $\mathbf{S} = \text{diag}\{S_2, S_2\}$ and sign $\stackrel{A}{\sim}$ represents ``asymptotically distributed as."

First, we show that with assumptions (a), (b) and (d), we have as $N' \rightarrow \infty$, $\sqrt{N'} (\boldsymbol{\hat{\beta}_M(k)}-\boldsymbol{\beta_0}-\theta_{N'}) \stackrel{A}{\sim} N(\mathbf{0},\mathbf{\Gamma_{N'}})$, where $\theta_{N'} := \mathbf{Q}^{-1}_{N'}\mathbf{b}_{N'}$ and $\mathbf{\Gamma_{N'}} := \mathbf{Q}^{-1}_{N'}\mathbf{W}_{N'}\mathbf{Q}^{-1}_{N'}$. Definitions of $\boldsymbol{\beta_0}$, $\mathbf{Q}_{N'}$, $\mathbf{W}_{N'}$ and $\mathbf{b}_{N'}$ are omitted here, which can be found in Appendix I of \cite{li2010nonlinear} where the asymptotic distribution of $\mathbf{L}_p$-norm periodogram is studied for $p \in (1,2)$. In the following, we drop the notation dependence of $j$ for simplicity.

% The key step is to reformulate the optimization problem~(11) for $\boldsymbol{\beta}$ to $\boldsymbol{\delta}$, and split the objective function as two parts, i.e., one in $L_2$ and one in $L_1$ form. With additional assumption $n_1 \ge n_2^2$, we can show that $L_1$ part in the objective function is asymptotically close to $o_P(1)$, where $o_P(1)$ represents a term that approaches zero in probability. As a result, the remaining $L_2$ part can be approximated by a quadratic function and an asymptotically Gaussian term using the similar argument as in \cite{li2010nonlinear}. 
To obtain the coefficient $\hat{\boldsymbol{\beta}}_M(k)$, we need to solve the problem \eqref{eq:huber-opt}. 
Define $t^1 := \{t||x_t - \boldsymbol{\phi}_t\boldsymbol{\beta}|\le \zeta\}$, $t^2 := \{t||x_t - \boldsymbol{\phi}_t\boldsymbol{\beta}|> \zeta\}$, and denote $\boldsymbol{\delta} := \sqrt{N'}(\boldsymbol{\beta}-\boldsymbol{\beta}_0)$, $v_t(\boldsymbol{\delta}):=\boldsymbol{\phi}_t\boldsymbol{\delta}/\sqrt{N'}$. Let the total error to model ${x_t}$ be $U_t:= x_t - \boldsymbol{\phi}_t\boldsymbol{\beta}_0$. Because $x_t = U_t +  \boldsymbol{\phi}_t\boldsymbol{\beta}_0$, it follows that $\hat{\boldsymbol{\beta}}_M(k)$ also minimizes the following $Z_{N'}(\boldsymbol{\delta})$ as:
% \begin{equation} % \begin{multline} equation
%     Z_{N'}(\boldsymbol{\delta}) = \frac{1}{2}\sum_{t\in t^1} \left(|U_t - v_t(\boldsymbol{\delta})|^2 - |U_t|^2\right) + \\ \frac{1}{2}\sum_{t\in t^2} \left(|U_t - v_t(\boldsymbol{\delta})| - |U_t|\right).
% \end{equation}\label{eq:reformualte-opt}
\begin{equation}\notag
\resizebox{\hsize}{!}{$\displaystyle
    Z_{N'}(\boldsymbol{\delta}) \!=\! \frac{1}{2}\!\sum_{t\in t^1} \!\left(|U_t - v_t(\boldsymbol{\delta})|^2 - |U_t|^2\right) +  \frac{1}{2}\!\sum_{t\in t^2} \!\left(|U_t - v_t(\boldsymbol{\delta})| - |U_t|\right).
$}
\end{equation}\label{eq:reformualte-opt}
We now want to show that $Z_{N'}(\boldsymbol{\delta})$ can be approximated by a quadratic function as
$
    Z_{N'}(\boldsymbol{\delta}) = \Tilde{Z}_{N'}(\boldsymbol{\delta}) + o_P(1)
$
for fixed $\boldsymbol{\delta}$. Based on Lemma 2.8 in \cite{arcones2001asymptotic}, since the result (vii) in Lemma 2.8 holds for both $p=1$ and 2, we are able to show that
\begin{multline}\label{eq:z_n_prime}
    \Tilde{Z}_{N'}(\boldsymbol{\delta}) = \sum\nolimits_{t\in t^1} \{-g_2(U_t)v_t + \frac{1}{2}h_2(U_t)v_t^2 + r_2(U_t,v_t)\} + \\
    \frac{1}{2}\sum\nolimits_{t\in t^2} \{-g_1(U_t)v_t + \frac{1}{2}h_1(U_t)v_t^2 + r_1(U_t,v_t))\},
\end{multline}
where 
$
r_i(u,v) := \text{min}\{|u|^{i-3}|v|^3,|u|^{i-2}|v|^2\}, i = 1,2$, and $h_p(x) := (p-1)|x|^{p-2}$.
Similar to \cite{li2010nonlinear}, we can rewrite Eq.~\eqref{eq:z_n_prime} as 
\begin{equation}
\Tilde{Z}_{N'}(\boldsymbol{\delta}) = T_{1N'}+T_{2N'}+T_{3N'}, 
\end{equation}
where $T_{1N'}:= -\sum_{t\in t^1} g_2(U_t)v_t$, $T_{2N'}:= \frac{1}{2} \sum_{t\in t^1} h_2(U_t)v_t^2$, and $T_{3N'} := \sum_{t\in t^1} r_2(U_t,v_t) + \frac{1}{2}\sum_{t\in t^2} (-g_1(U_t)v + r_1(U_t,v_t)))$.
The goal here is to assert that $Z_{N'}(\boldsymbol{\delta}) - T_{1N'}-(1/2)\boldsymbol{\delta}^T\mathbf{Q}_{N'}\boldsymbol{\delta} = o_P(1)$, and $T_{1N'} = -\boldsymbol{\delta}\boldsymbol{\zeta}_{N'} \stackrel{A}{\sim} N(-\sqrt{N'}\boldsymbol{\delta}^T\mathbf{b}_{N'},\boldsymbol{\delta}^T\mathbf{W_{N'}}\boldsymbol{\delta})$. Then the result as $N' \rightarrow \infty$, $\sqrt{N'} \boldsymbol{\hat{\beta}_M(k)} \stackrel{A}{\sim} N(\mathbf{0},2m_2\mathbf{S})$ follows directly.  
Since steps to show $T_{1N'}$ is asymptotically Gaussian and $T_{2N'}$ is approximated by a quadratic function is similar to Appendix I of \cite{li2010nonlinear}, we omit them here and focus on proving that $T_{3N'}$ is asymptotically negligible. 

To prove $\sum_{t\in t^1} r_2(U_t,v_t)+ \frac{1}{2}\sum_{t\in t^2} r_1(U_t,v_t)$ asymptotically goes to $o_P(1)$, we borrow the upper bounds that have been derived for $r_i(u,v), i=1,2$ by \cite{li2010nonlinear}, i.e., with $f_0 := \text{sup} f(x)$, then we have
\begin{align}\label{eq:Er1}
    E\{r_1(U_t,v_t)\} \!\le\!  f_0 v_t^2 \left( \!\int_{|x| \le |v_t|}^{|x-v_t| \ge \zeta}  \!  \! \!\frac{1}{|x|} dx
    +  v_t  \! \! \!\int_{|x| \ge |v_t|}^{|x-v_t| \ge \zeta} \!\! \! \! \frac{1}{|x|^2} dx \right),
\end{align}
\begin{align}\label{eq:Er2}
    E\{r_2(U_t,v_t)\}  \le   f_0 v_t^2  \left( \zeta
    +  v_t \int_{|x| \ge |v_t|}^{ |x-v_t| \le \zeta}   \frac{1}{|x|} dx \right).
\end{align}
% \begin{align}\label{eq:Er2}
%     E\{r_2(U_t,v_t)\} \!\le\!  f_0 v_t^2  \left( \! \!\int_{|x| \le |v_t|}^{ |x-v_t| \le \zeta} \!  \! \! |x|^{0} dx
%     +  v_t  \!  \! \!\int_{|x| \ge |v_t|}^{ |x-v_t| \le \zeta}  \!  \! \! \frac{1}{|x|} dx \right).
% \end{align}
% \begin{align}\label{eq:Er1}
%     E\{r_1(U_t,v_t)\} \le v_t^2 \cdot ( f_0 \int_{|x| \le |v_t| \& |x-v_t| \ge \zeta} |x|^{-1} dx \nonumber \\
%     + f_0 v_t \int_{|x| \ge |v_t| \& |x-v_t| \ge \zeta} x^{-2} dx).
% \end{align}
% \begin{align}\label{eq:Er2}
%     E\{r_2(U_t,v_t)\} \le v_t^2 \cdot ( f_0 \int_{|x| \le |v_t| \& |x-v_t| \le \zeta} |x|^{0} dx \nonumber \\
%     + f_0 v_t \int_{|x| \ge |v_t| \& |x-v_t| \le \zeta} |x|^{-1} dx ).
% \end{align}
Given our piece-wise nature of Huber loss, it is easy to show that for both $p = 1$ and $p=2$, the terms in parenthesis of~\eqref{eq:Er1} and~\eqref{eq:Er2} are finite with closed-form because the use of $\zeta$ with Huber loss function. Therefore, $E\{r_i(U_t,v_t)\}$ is bounded by $o_P(N'^{-1})$.
Furthermore, under assumption (d) that $n_1 \ge n_2^2$, we have $\sum_{t\in t^2} -g_1(u)v$ in $T_{3N'}$ is $o_P(n_2/\sqrt{n_1+n_2}) \approx o_P(1)$. Therefore, we proof that $T_{3N'}$ is asymptotically negligible. Finally, Proposition~3.3 is a direct result under assumption (c) when $\boldsymbol{\beta}_{0} = \theta_{N'} =0$.
\end{proof}
% \begin{proof}
% {\color{black}
% The main step of the proof is to obtain $\sqrt{N'} \boldsymbol{\hat{\beta}_M(k)} \stackrel{A}{\sim} N(\mathbf{0},2m_2\mathbf{S})$ as $N' \rightarrow \infty$, where $\boldsymbol{\hat{\beta}_M(k)}$ is defined as $\boldsymbol{\hat{\beta}_M(k)} := \text{arg min}_{\boldsymbol{\beta} \in \boldsymbol{R}^2} \sum\nolimits_{t=1}^{N'} \gamma( \boldsymbol{\phi}_t \boldsymbol{\beta} -  x_t )$ and $\mathbf{S} = \text{diag}\{S_2, S_2\}$. Then, under the definition of Huber-periodogram in Eq.~\eqref{eq:Periodogram_final}, it can be readily obtained that $P_k^M$ is asymptotically distributed as scaled $\chi^2$-distribution with 2 degrees of freedom as $P_k^M \stackrel{A}{\sim} (1/2)m_2S_2\chi_2^2$. 
% Due to the space limitation, we leave the detailed proof in the Section 3.4.1 of the extended technique report online\footnote{{\color{black}https://arxiv.org/abs/2002.09535}}.  
% }
% \end{proof}
Proposition \ref{prop:HuberPeriodogram_dist} indicates that the Huber-periodogram behaves similarly to the vanilla periodogram as $n\rightarrow \infty$. Therefore, the Fisher's test based on Huber-periodogram can also be utilized to detection periodicity. 

Furthermore, for the $j$th level data $\mathbf{x}_j$ from $\mathbf{w}_j$, we only calculate $P_k^M$ at frequency indices $[{N'}/{2^{j+1}},{N'}/{2^{j}}]$ based on Eq.~\eqref{eq:Periodogram_final} since the possible period length is within $[2^j, 2^{j+1}]$ at $j$th level, while using Eq.~\eqref{eq:Periodogram} to approximate $P_k^M$ at the rest frequency indices to speed up the computation.

\subsubsection{Robust Huber-Periodogram based ACF for Validating Periodicity Candidates}\label{sec:RobustACF}
After obtaining  period candidate from the robust Fisher's test for each wavelet coefficient, we next validate each candidate and improve its accuracy by using ACF. This step is necessary since periodogram has limited resolution and spectral leakage~\cite{cPD_vlachos2005Autoperiod}, which makes the candidate from Fisher's test not accurate.

For the ACF of the time series from wavelet coefficient  $w_{j,t}$ (denote as $w_t$ for simple notation), the normalized estimation~\cite{TSanalysisBook_box2015time} is
\begin{equation}\notag %\label{eq:ACFfun} 
ACF(t) = \frac{1}{(N-t)\delta_{w}^2}\!\sum\nolimits_{n=0}^{N-t-1} \!w_{n} w_{n+t}, ~~ t = 0,1,\cdots,N-1,
\end{equation}
where $\delta_{w}$ is the sample variance of $w_t$.
However, this conventional ACF is not robust to outliers and has $O(N^2)$ complexity. Instead, we propose to utilize the output of Huber-periodogram to obtain robust ACF with $O(NlogN)$ complexity. Specifically, since the time series is real-valued data, we can have the full-range periodogram
$$
\bar{P}_k =
\begin{cases} 
P_k^M        & k=0,1, \cdots, N -\! 1 \\
{\left(\sum_{k=0}^{N-1} x_{2k} - x_{2k+1}\right)^2} \Big/ {N'}    & k = N \\
P_{N'-k}^M       & k=N \!+ \!1, \cdots, N'\!-\!1  \\
\end{cases}
$$
Then, based on Wiener-Khinchin theorem \cite{Wiener1930}, we obtain the robust ACF (denote as Huber-ACF) as
\begin{equation}\label{eq:rboustACF} 
HuberACF(t) = \frac{p_t}{(N-t)p_0}, t = 0,1,\cdots,N-1
\end{equation}
where $p_t$ is the IDFT as
% \begin{equation}\label{eq:IDFT} 
$p_t = \text{IDFT}\{\bar{P}_k\} = \frac{1}{\sqrt{N'}}\sum_{k=0}^{N'-1} \bar{P}_k e^{i2 \pi kt/N'}$. 
% \end{equation}
Since we aim to detect single dominant periodicity in each level of wavelet coefficient, we summarize the peaks of the Huber-ACF through 
peak detection~\cite{palshikar2009simple}. Then, we calculate the median distance of those peaks whose heights exceed the predefined threshold. Furthermore, based on the resolution of periodogram, i.e., the peak value of $P_k^M$ at index $k$ corresponds to period length in the range $[\frac{N}{k}, \frac{N}{k-1})$, the median distance of Huber-ACF peaks is the final period length only if it locates in the range of 
\begin{equation}\notag%\label{eq:ACFrange} 
R_k = \left[\frac{1}{2}\left(\frac{N}{k+1} + \frac{N}{k}\right)-1, \cdots, \frac{1}{2}\left(\frac{N}{k} + \frac{N}{k-1}\right)+1\right].
\end{equation}
We denote the above described procedure as Huber-ACF-Med. By summarizing all the periods from the Huber-ACF-Med at different level of wavelet coefficients, we obtain the final periods of the original time series.

\section{Experiments and Discussions}

In this section, we evaluate and discuss the proposed RobustPeriod algorithm with other state-of-the-art periodicity detection algorithms on both synthetic and real-world datasets.
% , and discuss how each component in RobustPeriod contributes to the final accurate detection.

% In this section, we study the proposed RobustPeriod algorithm empirically in comparison with other state-of-the-art periodicity detection algorithms on both synthetic and public real-world benchmark datasets. We also investigate how each component in RobustPeriod contributes to the final accurate detection.
% We conduct experiments on synthetic datasets and public datasets to demonstrate the effectiveness of the proposed RobustPeriod algorithm for periodicity detection and estimation.

% \vspace{-0.2cm}
\subsection{Baseline Algorithms and Datasets}

\subsubsection{Existing Algorithms}
We consider three single-periodicity detection algorithms.
1) {findFrequency}~\cite{hyndman2019package}: it is based on maximum value in frequency spectrum to estimate period length. Similar methods can be also found in search queries~\cite{vlachos2004identifying,drutsa2017periodicity} and cloud workload modelling~\cite{cortez2017resource}.
2) {SAZED$_{maj}$} and 3)  {SAZED$_{opt}$}~\cite{Toller2019}: these two methods adopt majority vote and optimal ensemble, respectively. We also consider three multi-periodicity detection algorithms.
4) {Siegel}~\cite{Siegel1980,walden1992asymptotic}: it is a periodogram based method by extending Fisher's test to support multiple periods detection. Similar methods can be found in cloud workload modelling~\cite{Telescope2020ICDE}.
5) {AUTOPERIOD}~\cite{cPD_vlachos2005Autoperiod,Mitsa:2010:TDM:1809755}: it is a combination method based on periodogram and ACF.
6) {Wavelet-Fisher}~\cite{Abdullah:testing:periodicity}: it adopts DWT to decouple multiple periodicities and then use Fisher's test to detect single periodicity at each level.

As the trend component may bias the periodicity detection results significantly, we apply HP filter to remove the trend component for all algorithms for a fair comparison in our experiments.

\subsubsection{Synthetic Datasets}
\begin{figure}[!t]
    \centering
    \vspace{-0.2cm}
    \subfigure[The generation of synthetic data with 3 periods.]{
        \includegraphics[width=0.95\linewidth]{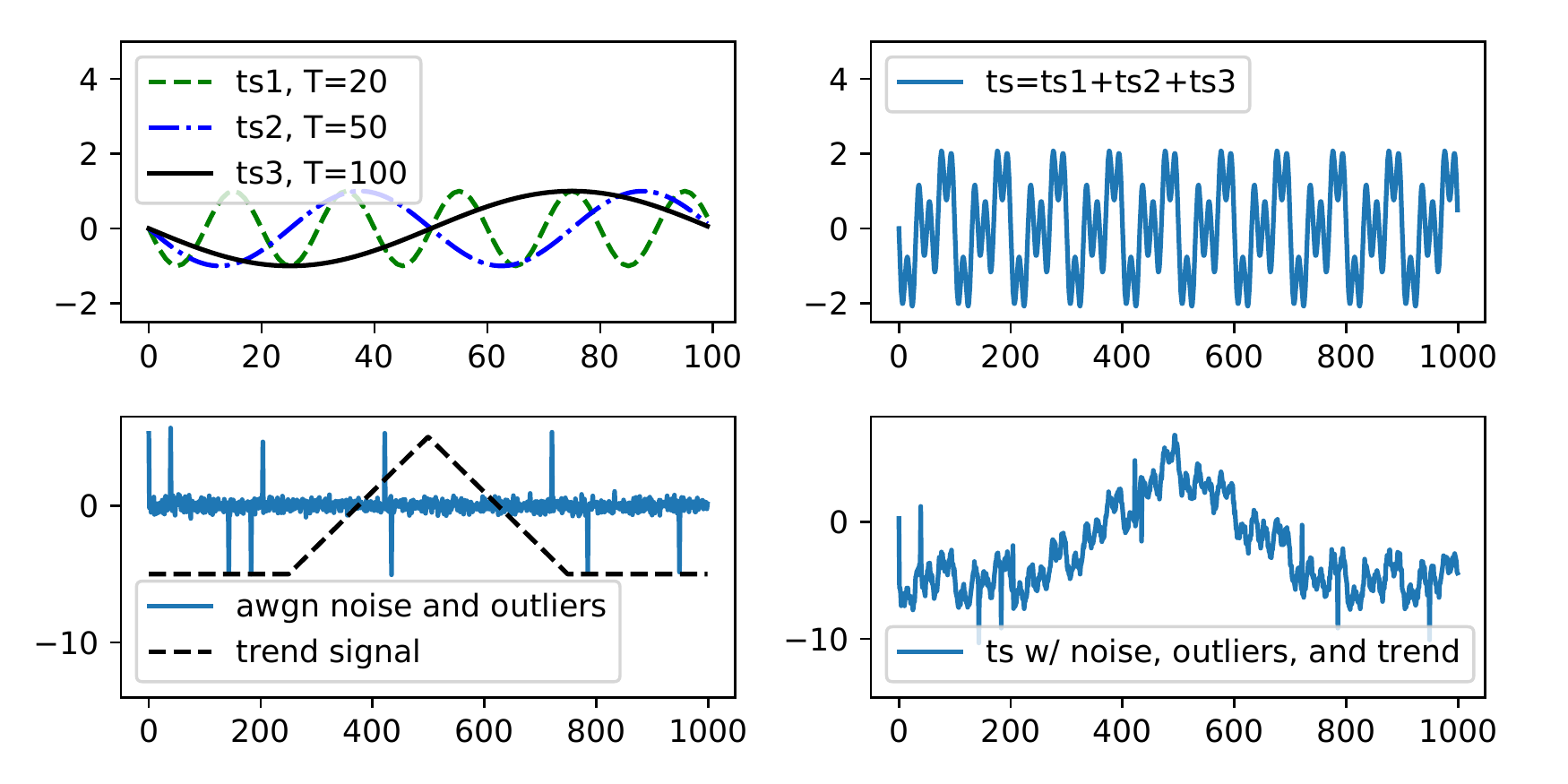}
        \label{fig:simdata}
        }
        \\ \vspace{-0.3cm}
    \subfigure[Representative public datasets from CRAN and Yahoo.]{
        \includegraphics[width=0.95\linewidth]{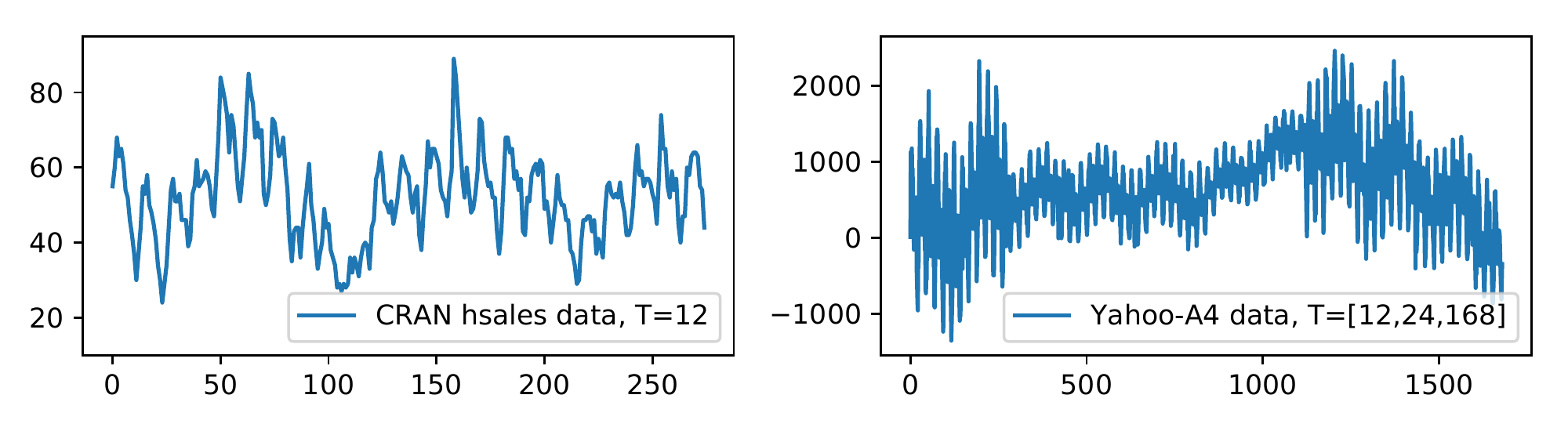}
        \label{fig:pubdata}
        } \vspace{-0.5cm}
    \caption{Synthetic and public periodic time series data.}
    \vspace{-0.3cm}
    % \label{fig:pubdata}
\end{figure}

\begin{figure}[t]
\centering
    \vspace{-0.2cm}
    \includegraphics[width=0.95\linewidth]{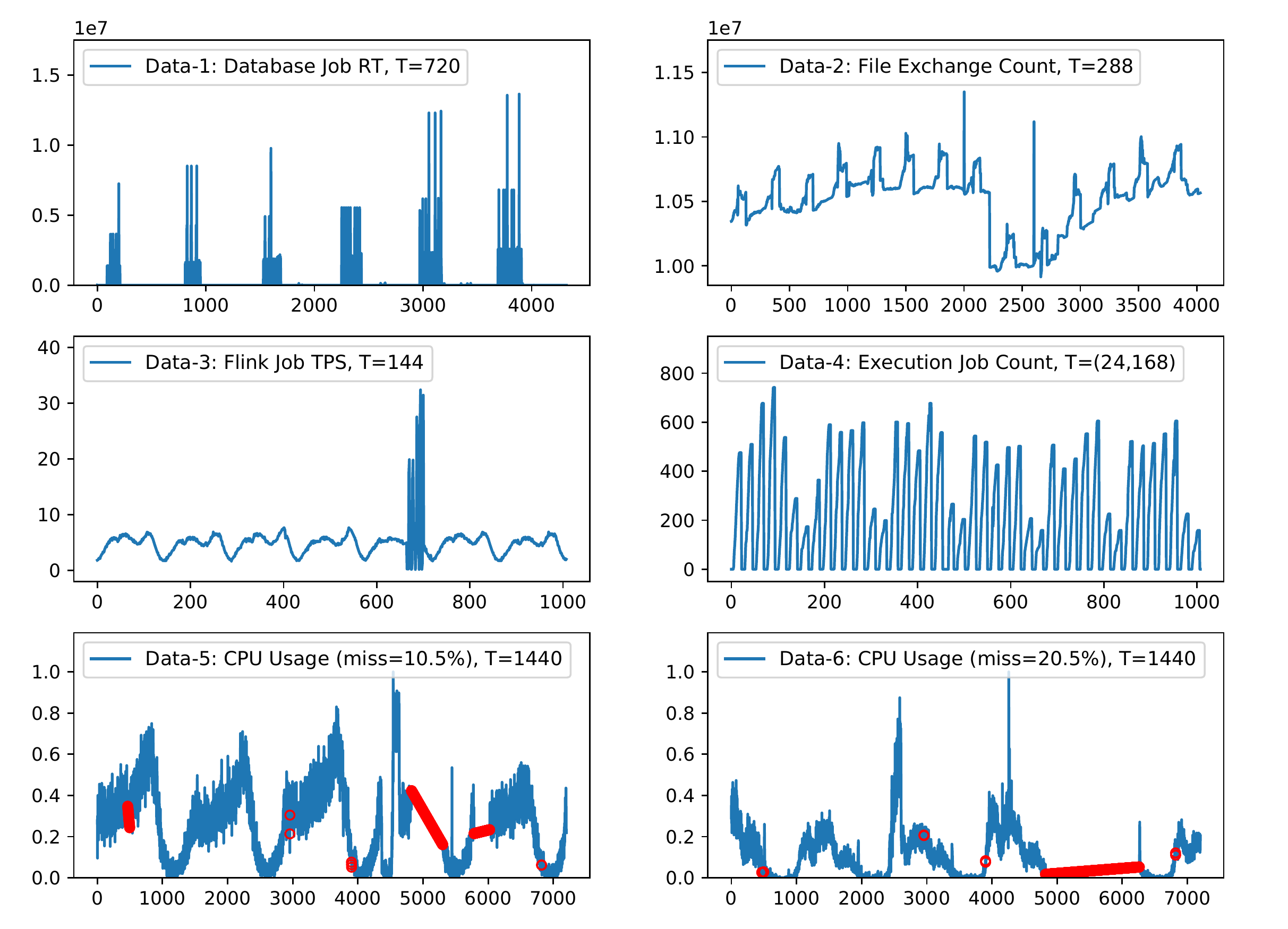}
    \vspace{-0.4cm}
    % \caption{{\color{red}{Monitoring datasets in cloud database/computing.}}}
    % \caption{Monitoring datasets in cloud database/computing.}
    \caption{Monitoring datasets from Alibaba cloud database/computing.}
    \vspace{-0.5cm}
    % \caption{Five representative real-world datasets: the first three datasets contain single period with length $12, 144, 288$, respectively; the last two datasets contain double periods with lengths as $(6,12), (48,336)$, respectively.}
\label{fig:realDatasets}
\end{figure}

% {\color{red}
We generate synthetic datasets under different conditions to quantitatively evaluate the performance of all periodicity detection algorithms, especially when the common challenging characteristics of time series (outliers, noise, trend change, etc.) for periodicity detection exhibit. Specifically, we generate both single-period and multi-period time series with noise, outliers, and changing trend. For the base periodic signal, we adopt sinusoidal wave to approximate the usual scenarios. Besides, we also adopt square-wave and triangle-wave signal to represent real-world non-sinusoidal cases, which are more challenging for periodicity detection algorithms. 
Note that all these challenging characteristics can be found in real-world time series as shown in Fig.~\ref{fig:pubdata} (public CRAN datasets) and Fig.~\ref{fig:realDatasets} (cloud database/computing datasets). Meanwhile, the extent of noises and the amount of outliers are generated by corresponding controllable parameters, which are used for algorithm evaluation under mild or severe conditions. Furthermore, the multi-period synthetic dataset is also utilized to illustrate how the proposed RobustPeriod algorithm can effectively and accurately detect multiple periodicities in challenging time series as shown in Fig.~\ref{fig:simdata} and Fig.~\ref{fig:tfDetail}.

For the detailed procedure of synthetic datasets, 
we first generate synthetic time series of length $1000$ with complex patterns, including $3$ periodic components, multiple outliers, changing trend, and noises. Specifically, we generate $3$ sinusoidal, square, or triangle waves with amplitude of $1$ and period lengths of $20, 50, 100$. Then we add a triangle signal with amplitude of $10$ as trend. We add Gaussian noise and outliers in different scenarios: mild condition (noise variance $\sigma_n^2=0.1$, outlier ratio $\eta=0.01$) and severe conditions ($\sigma_n^2=1$ or $2$ and $\eta=0.1$ or $0.2$). For single-period case, we only pick the periodic component with period $100$. In all experiments, we randomly generate 1000 time series for evaluation. One synthetic sin-wave data with mild condition is illustrated in Fig.~\ref{fig:simdata}.

% \vspace{-0.3cm}
\subsubsection{Public Datasets}
We use public single-period data from CRAN dataset as in~\cite{Toller2019} which contains 82 real-world time series from a wide variety of domains, such as retail sales, electricity usage, pollution levels, etc. The length of these time series ranges from 16 to 3024, and their period length ranges from 2 to 52. 
We adopt public multiple-period data from Yahoo's webscope S5 datasets~\cite{yahooS5,Laptev_2015} which includes the Yahoo-A3 and Yahoo-A4. These datasets contains 200 time series, and each time series contains 1680 points with 3 period lengths of 12, 24, and 168. The representative data from CRAN and Yahoo are illustrated in Fig.~\ref{fig:pubdata}. 
% {\color{red} %for major added contents in revised version
% (TODO:QSW). add 2 more data if have time
% }

% \vspace{-0.3cm}
\subsubsection{Cloud Monitoring Datasets}
% For periodicity detection demonstration, we also select 4 representative real-world datasets from the monitoring system in a leading cloud computing company shown in Fig.~\ref{fig:realDatasets}. These datasets are used for workload forecasting, anomaly detection, and auto-scaling of cloud database/computing. The first 3 datasets are with a single period (daily pattern) while the last dataset has double periods (daily and weekly), where the period lengths are listed in Fig.~\ref{fig:realDatasets}. Due to different recording time intervals, the period length of daily pattern can be different. Note that there are also periodic/trend pattern changes, lots of noise and outliers in these datasets. 

For periodicity detection demonstration, we also select 6 representative challenging real-world datasets from the monitoring system of Alibaba Cloud as shown in Fig.~\ref{fig:realDatasets}. These datasets are used for workload forecasting, anomaly detection, and auto-scaling of cloud database/computing. It can be observed that these challenging datasets contain periodic/trend pattern changes, lots of noise and outliers, even block missing data. The first 3 datasets are with a single period (daily pattern) while the 4th dataset has double periods (daily and weekly). 
% {\color{red}
Note that the last 2 datasets (daily pattern) contain lots of missing data, which are linearly interpolated (marked by red circles) before sent to different periodicity detection algorithms. The true period lengths of these datasets are listed in Fig. ~\ref{fig:realDatasets}. Note that the period length of daily pattern may be different due to different recording time intervals (varying from 1 minute to 1 hour).  
% }
% The length of these dataset is 5 days with sampling resolution of 10 minutes, where their true period lengths are 144 (one-day), 72 (half-day), and 144 respectively.

% \vspace{-0.2cm}
\subsection{Comparisons with Existing Algorithms}

\subsubsection{Single-Periodicity Detection}
We summarize the detection precision of single-periodicity detection algorithms on both synthetic and public CRAN datasets in Table~\ref{tab:single_period_det}, where $\pm2\%$ indicates that detection is considered correct if the detected period length is within a $2\%$ tolerance interval around the ground truth while $\pm0\%$ indicates that we only consider exactly match. As the CRAN data contains both simple and complex periodic time series, the difference is not significant between different algorithms. 
For synthetic data, findFrequency cannot find the correct periodicity. The reason is that findFrequency fits an autoregression model for spectral density estimation when finding periodicity, while the added outliers make the autoregression model not accurate. In all cases, SAZED$_{opt}$ outperforms SAZED$_{maj}$ since the former uses its proposed optimal ensemble method while the later adopts a majority vote. Overall, RobustPeriod achieves the best performance.

\begin{table}[t] % \begin{table*}[t]
\centering
\footnotesize
% \vspace{-0.2cm}
\caption{Precision comparisons of single-period detection algorithms on synthetic sin-wave data and public CRAN data.}
% \caption{Precision of different single-periodicity detection algorithms on synthetic data with different noise variance $\sigma^2_n$ and outlier ratio $\eta$, as well as the real-world dataset from CRAN.}
\vspace{-0.3cm}
\label{tab:single_period_det}
\begin{tabular}{c|c|c|c|c|c|c}
\hline
\multirow{3}{*}{Algorithms} & \multicolumn{4}{c|}{Synthetic Sin Data}  & \multicolumn{2}{c}{\multirow{2}{*}{CRAN Data}} \\ \cline{2-5}
                  & \multicolumn{2}{c|}{\!$\sigma^2_n$ \!=\! 0.1, $\eta$ \!=\! 0.01\!} & \multicolumn{2}{c|}{\!$\sigma^2_n$ \!=\! 2, $\eta$ \!=\! 0.2\!} & \multicolumn{2}{c}{}                  \\ \cline{2-7} 
 &  $\pm$0\% & $\pm$2\% & $\pm$0\%   &  $\pm$2\% & $\pm$0\% & $\pm$2\% \\ \hline\hline
findFrequency   &  0 & 0&        0 &0  & 0.44 & 0.44\\ \hline
% SAZED$_{maj}$      & 0 &  0          & 0  & 0.32   &  0.49 & 0.49 \\ \hline
%% do new exp (in futrue)
SAZED$_{maj}$      & 0 &  0.32          & 0  & 0   &  0.49 & 0.49 \\ \hline
SAZED$_{opt}$      & 0  & 0.96       & 0  & 0.54   & 0.55  & 0.56  \\ \hline
\textbf{RobustPeriod} &  \textbf{0.83}  & \textbf{1.0}    & \textbf{0.44}  & \textbf{0.98}   &  \textbf{0.60}  & \textbf{0.61}  \\ \hline        
\end{tabular}
\vspace{-0.4cm}
\end{table}

%     \item \textbf{findFrequency}~\cite{hyndman2019package}: It first removes a fitted linear trend, then the spectral density function is estimated from the best fitting autoregressive model (based on the AIC). Finally, the reciprocal of the
%     maximum in the spectral density function is estimated as the period length. 
%     \item \textbf{SAZED$_{maj}$}~\cite{Toller2019}: It estimates a time series’ season length by computing six different estimates and taking a majority vote.
%     \item \textbf{SAZED$_{opt}$}~\cite{Toller2019}: It estimates a time series’ season length by combining three different estimates computed on an input time series and its ten-fold self-composed autocorrelation.

\subsubsection{Multi-Periodicity Detection}

In multi-periodicity detection, we use F1 score to evaluate different algorithms as multiple periodicities are compared. The F1 scores of different algorithms on both synthetic sin-wave datasets and Yahoo datasets are summarized in Table~\ref{tab:multi_period_det}. For synthetic sin-wave data, Siegel algorithm has better performance than other existing algorithms, and also has relatively stable performance under $\pm0\%$ and $\pm2\%$ tolerance. While in Yahoo data, AUTOPERIOD has better performance than other existing algorithms. In both datasets, our RobustPeriod algorithm achieves the best performance.

\begin{table}[t] % \begin{table*}[t]
\centering
\footnotesize
% \tiny, \scriptsize, \footnotesize, \small, \normalsize, 
% \caption{F1 scores of different Multi-periodicity detection algorithms on synthetic data with different noise variance $\sigma^2_n$ and outlier ratio $\eta$, as well as the public dataset from Yahoo.}
\caption{F1 score comparisons of multi-period detection algorithms on synthetic sin-wave data and public Yahoo data.}
\vspace{-0.3cm}
\label{tab:multi_period_det}
\begin{tabular}{c|c|c|c|c|c|c|c|c}
\hline
\multirow{3}{*}{Algorithms} & \multicolumn{4}{c|}{Synthetic Sin Data}  & \multicolumn{2}{c|}{\multirow{2}{*}{Yahoo-A3}} & \multicolumn{2}{c}{\multirow{2}{*}{Yahoo-A4}} \\ \cline{2-5}
                  & \multicolumn{2}{c|}{\!\!\!$\sigma^2_n$ \!=\! 0.1, \!$\eta$ \!=\! 0.01\!\!} & \multicolumn{2}{c|}{\!\!$\sigma^2_n$ \!=\! 1, \!$\eta$ \!=\! 0.1\!\!} & \multicolumn{2}{c|}{}  & \multicolumn{2}{c}{}                 \\ \cline{2-9} 
 &  $\pm$0\% & $\pm$2\% & $\pm$0\%   &  $\pm$2\% & $\pm$0\% & $\pm$2\% & $\pm$0\% & $\pm$2\% \\ \hline\hline
 %% do new exp (in futrue for seigel)
Siegel                     & 0.79 & 0.80    &  0.67       & 0.68            & 0.75 & 0.75& 0.75 & 0.75    \\ \hline
\!AUTOPERIOD\!           &  0.25       & 0.51        & 0.17 &0.42           & 0.80  & 0.80 & 0.80 & 0.80   \\ \hline
\!Wavelet-Fisher\!       &  0.50        & 0.75        & 0.48  &0.72          & 0.50  & 0.76  & 0.49& 0.73  \\ \hline
\!\textbf{RobustPeriod}\!&\textbf{0.99}&\textbf{0.99}& \textbf{0.92}  &\textbf{0.98} &  \textbf{0.82} & \textbf{0.82}& \textbf{0.83} &\textbf{0.84}\\ \hline        
\end{tabular}
\vspace{-0.2cm}
\end{table}

% {\color{red} %for major added contents in revised version
For synthetic data, besides sin-wave based periodic time series, we also compare the performance of 3-periodic square-wave and triangle-wave datasets under noise variance $\sigma_n^2=0.1$ and outlier ratio $\eta=0.01$, which are adopted to represent non-sinusoidal data in more challenging scenarios. Table~\ref{tab:multi_period_det_square} summarizes the F1 scores of different periodicity detection algorithms. It can be observed that most algorithms cannot handle the non-sinusoidal data properly and achieve worse performance. In contrast, our algorithm still achieves desirable results and exhibits much better performance than others. 
% }

\begin{table}[t] % \begin{table*}[t]
\centering
\footnotesize
% \tiny, \scriptsize, \footnotesize, \small, \normalsize, 
% \caption{F1 scores of different Multi-periodicity detection algorithms on synthetic data with different noise variance $\sigma^2_n$ and outlier ratio $\eta$, as well as the public dataset from Yahoo.}
\caption{{\color{black}{F1 score comparisons of multi-period detection algorithms on synthetic square- and triangle-wave datasets.}}}
\vspace{-0.3cm}
\label{tab:multi_period_det_square}
{\color{black}
\begin{tabular}{c|c|c|c|c}
\hline
\multirow{2}{*}{Algorithms}  & \multicolumn{2}{|c}{Synthetic Square} & \multicolumn{2}{|c}{Synthetic Triangle}                 \\ \cline{2-5} 
                %   & \multicolumn{2}{c|}{ $\sigma^2_n$  =0.1, $\eta$= 0.01} & \multicolumn{2}{|c}{ $\sigma^2_n$  =0.1, $\eta$= 0.01}                  \\ \cline{2-5} 
 &  $\pm$0\% & $\pm$2\% & $\pm$0\%   &  $\pm$2\%  \\ \hline\hline
 %% do new exp (in futrue for seigel)
Siegel                   &  0.53       & 0.53        & 0.55  &  0.55            \\ \hline
\!AUTOPERIOD\!           &  0.60       & 0.60        &  0.19   & 0.42            \\ \hline
\!Wavelet-Fisher\!       &  0.44       & 0.67        &  0.45   & 0.67           \\ \hline
\!\textbf{RobustPeriod}\!&\textbf{0.95}&\textbf{0.95}& \textbf{0.88}  &\textbf{0.99}  \\ \hline     
\end{tabular}
}
\vspace{-0.2cm}
\end{table}

 %for major added contents in revised version
\subsubsection{Real-World Representative Datasets}\label{sec:real_monitoring_data}
We compare the performance of 6 representative challenging real-world datasets from Alibaba Cloud as shown in Fig.~\ref{fig:realDatasets}.
The detection results are summarized in Table~\ref{tab:realData_diff_algs}. 
It can be observed that many existing algorithms may have false positive results. Also, due to the challenging patterns in the datasets, the existing methods often cannot obtain the accurate period length. In contrast, the proposed RobustPeriod achieves the best results in all 6 challenging datasets.
{\color{black}
In particular, other algorithms fail on Data-5 and Data-6 datasets due to its complex patterns, including 10\% to 20\% missing data (which are linearly interpolated before periodicity detection), severe noise and outliers. Even in these two extremely challenging scenarios, the detection error of the proposed RobustPeriod algorithm is still less than 1\% without false positive. In fact, these small errors of detected periodic length can be easily corrected in practice by domain knowledge. 
}

\begin{table}[t]%[!htbp]%[!htb]
\footnotesize 
\centering
\caption{{\color{black}{Comparisons of periodicity detection on 6 real-world datasets from Alibaba cloud database/computing.}}}
\vspace{-0.3cm}
\label{tab:realData_diff_algs}
{\color{black}
\begin{tabular}{c|c|c|c}
\hline
\multirow{2}{*}{Algorithms}       & Data-1, T=720       & Data-2, T=288     & Data-3, T=144       \\
 & \!Database RT\!  & \!\! File Exchange \!\! &\! Flink TPS  \!    \\\hline \hline
Siegel       & (655,769,...)   & (288,576,...)     & (141,144)       \\ \hline
AUTOPERIOD       & (353,241,9)   & (288,439,...)     & (68,141)      \\ \hline
Wavelet-Fisher   & (372,745,...) & (282,585,...)     & (73,146)   \\ \hline
\textbf{RobustPeriod}  & \textbf{721}   & \textbf{288}               & \textbf{144}             \\ \hline
% Ground truth           & {720}   & 288               & 144           \\ \hline
\end{tabular}

\begin{tabular}{c|c|c|c}
\hline
\multirow{2}{*}{Algorithms}   & Data-4, T=(24,168)   & Data-5, T=1440       & Data-6, T=1440          \\
& \!Job Count  \!  & \!\! CPU Usage \!\! &\! CPU Usage  \!      \\\hline \hline
Siegel       & (24,168)             & (1459,2597,...)     & (1575,1063,...)       \\ \hline
AUTOPERIOD      & (24,26)            & (1488,739,...)     & (366,2880,...)     \\ \hline
Wavelet-Fisher   & (12,24,...)       & (1489,712,...)     & (1489,364,...)   \\ \hline
\textbf{RobustPeriod}  & (\textbf{24},\textbf{168})   & \textbf{1431}     & \textbf{1426}    \\ \hline
% Ground truth          & (24,168)   & 288               & 144            \\ \hline
\end{tabular}

}
\vspace{-0.4cm}
\end{table}

\subsection{Ablation Studies and Discussion}

\subsubsection{Ablation Studies}
% \begin{itemize}
%     \item \textbf{Huber-ACF-Med}: It is based on the previous \textit{ACF-Med} by using Huber-periodogram and robust ACF.
%     \item \textbf{Huber-Fisher}: This algorithm has the same procedure as the \textit{Fisher} except that the vanilla periodogram is replaced by Huber-periodogram.
%     \item \textbf{Huber-Siegel-ACF}: This algorithm also adopt Huber-periodogram when finding multiple period candidates using Siegel's test. Then, the candidates are validated by checking if they are located near the peaks of ACF as in AUTOPERIOD.
%     \item \textbf{NR-RobustPeriod}: This one is non-robust version of RobustPeriod by using vanilla periodogram and ACF while the structure is the same as RobustPeriod.
% \end{itemize}

% (Ablation studies)
% {\color{red} %for major added contents in revised version
To further understand the contribution of each component in our RobustPeriod algorithm, we compare the performance of RobustPeriod with the following ablation revisions:
% 1) \textbf{Huber-ACF-Med}: It is based on the previous \textit{ACF-Med} by using Huber-periodogram and robust ACF;
1) \textbf{Huber-Fisher}: This algorithm replaces the vanilla periodogram in Fisher's test with Huber-periodogram;
%shares the same procedure as the \textit{Fisher} except that the vanilla periodogram is replaced by Huber-periodogram;
2) \textbf{Huber-Siegel-ACF}: This algorithm also adopts Huber-periodogram when finding multiple period candidates in Siegel's test. Then, the candidates are validated by checking if they are located near the peaks of ACF as in AUTOPERIOD;
3) \textbf{NR-RobustPeriod}: This one is the non-robust version of RobustPeriod by using vanilla wavelet variance, periodogram, and ACF while sharing the same procedure as RobustPeriod.

Table~\ref{tab:Ablation_new} summarizes the detailed periodicity detection results (precision, recall, and F1 score) of the aforementioned revisions on the synthetic sin-wave data under noise variance $\sigma_n^2=2$ and outlier ratio $\eta=0.2$. It can be observed that all ablation revisions have some performance degradation in comparison with RobustPeriod, and the proposed RobustPeriod algorithm achieves the best performance. 
% }
 
\begin{table}[h] % \begin{table*}[t]
% \footnotesize 
\centering
\small
%\normalsize
% \tiny, \scriptsize, \footnotesize, \small, \normalsize, \large, \Large, \LARGE, \huge, and \Huge.
% \vspace{-0.2cm}
\caption{Ablation studies of the proposed RobustPeriod on synthetic data.}
\vspace{-0.3cm}
\label{tab:Ablation_new}
% {\color{red}
\begin{tabular}{c|c|c|c|c|c|c}
\hline  %          &   
\multirow{2}{*}{Algorithms}    & \multicolumn{3}{|c}{ tolerance=$\pm$0\% } & \multicolumn{3}{|c}{  tolerance=$\pm$2\%    } \\ \cline{2-7}
%  &  pr & re & f1     & pr  & re & f1\\ \hline\hline
   &  pre & recall & f1      &  pre & recall & f1 \\ \hline\hline
Huber-Fisher  & \textbf{0.91}  &0.3   & 0.46                 &  0.89          & 0.3         & 0.45 \\  \hline
Huber-Siegel-ACF     & 0.09  & 0.28  & 0.13                 & 0.25          &0.55         & 0.31 \\ \hline
NR-RobustPeriod      & 0.71  & 0.6   & 0.64                & 0.96           &0.79         & 0.85 \\ \hline
\textbf{RobustPeriod}  & 0.76  & \textbf{0.7}  &\textbf{0.72}  & \textbf{0.98} & \textbf{0.91}& \textbf{0.93}  \\ \hline
\end{tabular}
% }
\vspace{-0.4cm}
\end{table}

% \vspace{-0.1cm}
\subsubsection{Effectiveness of MODWT Decomposition} \label{effect_MODWT_decomp}

\begin{figure}[!t]
    \centering
    \subfigure[Left to right: Wavelet coefficient, Huber-periodogram and ACF. The true period lengths 20, 50, 100 are correctly detected at level 4,5,6.]{
        \includegraphics[width=0.99\linewidth]{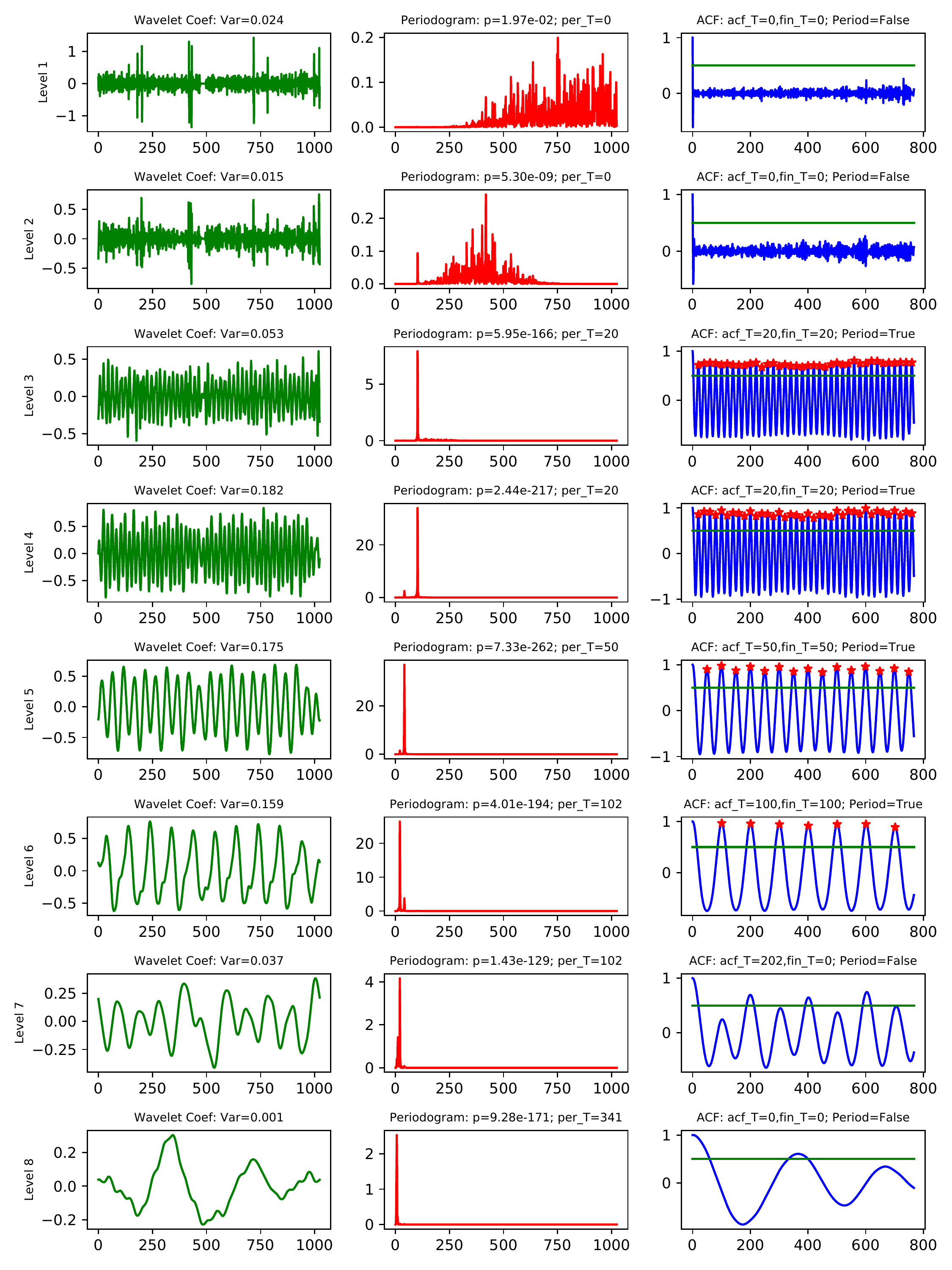}
        \label{fig:tfDetail:all_level}
        }
        \\ \vspace{-0.4cm}
    \subfigure[Wavelet variance: the true periodic components located at levels 4,5,6 also have largest wavelet variances.]{
        \includegraphics[width=0.95\linewidth]{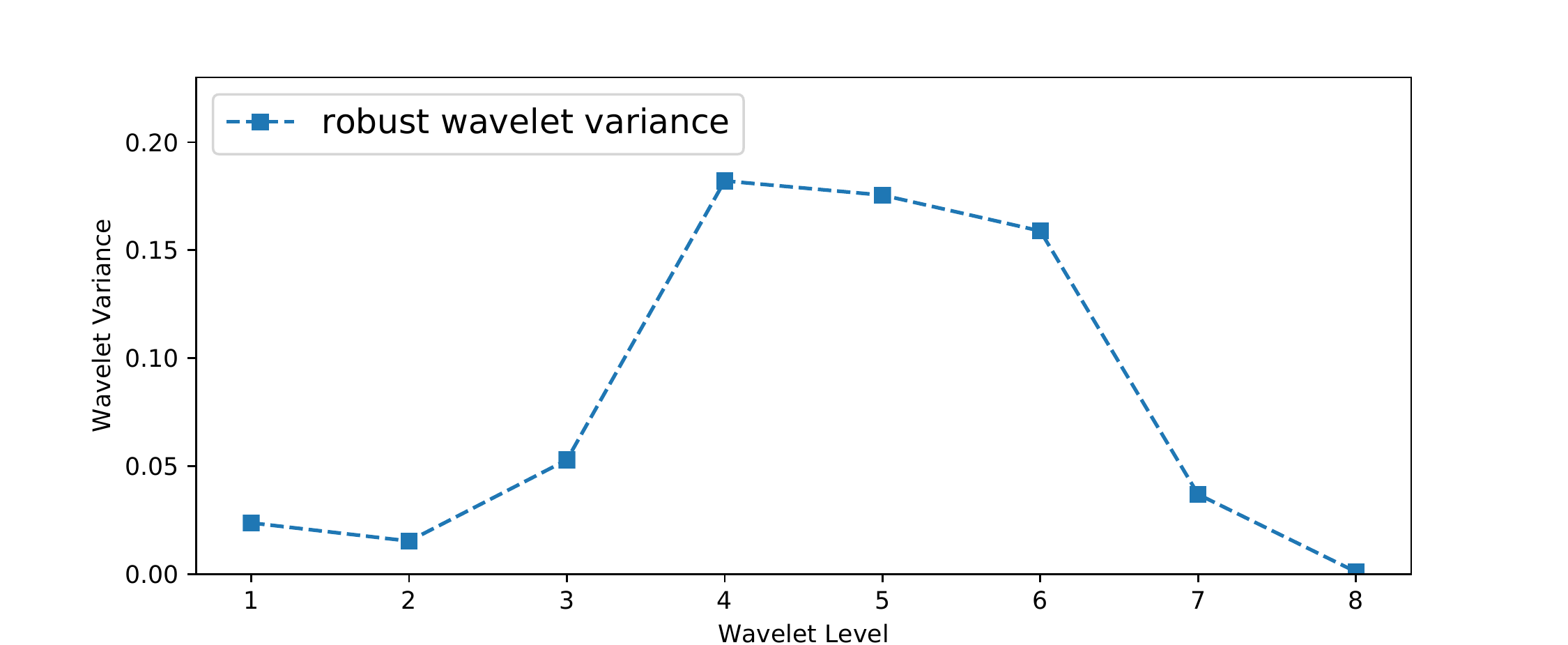}
        \label{fig:tfDetail:WV}
        } \vspace{-0.4cm}
    \caption{Intermediate results of the RobustPeriod.}
    \vspace{-0.5cm}
    \label{fig:tfDetail}
\end{figure}
%  at different level: the levels 4,5,6 have high wavelet variance, which also corresponds to the location of periodic components]

To further understand how RobustPeriod detects multiple periodicities, we plot the intermediate results in Fig.~\ref{fig:tfDetail:all_level} for the synthetic dataset from Fig.~\ref{fig:simdata}, where the first column is the wavelet coefficient, the second column is the Huber-periodogram, and the last column is the Huber ACF, and each row corresponds to a wavelet coefficient at a specific level. 
It can be observed that MODWT effectively decouples the interlaced periodicities. The Huber-periodogram and ACF effectively detect the periods of 20, 50, 100 at level 4, 5, 6, respectively. As a comparison, AUTOPERIOD cannot detect the period of $50$ as the vanilla ACF does not have peak near 50 (the vanilla ACF drops near 50 due to the strong periodicities of 20 and 100). Fig.~\ref{fig:tfDetail:WV} plots the wavelets variances at different levels. It is clear that largest wavelet variances correspond to strong periodic patterns at levels 4, 5, and 6.

%%%% in icdm
\begin{figure}[!t]
    \centering
    \subfigure[Time series in normal/abnormal cases ]{\includegraphics[width=0.23\textwidth]{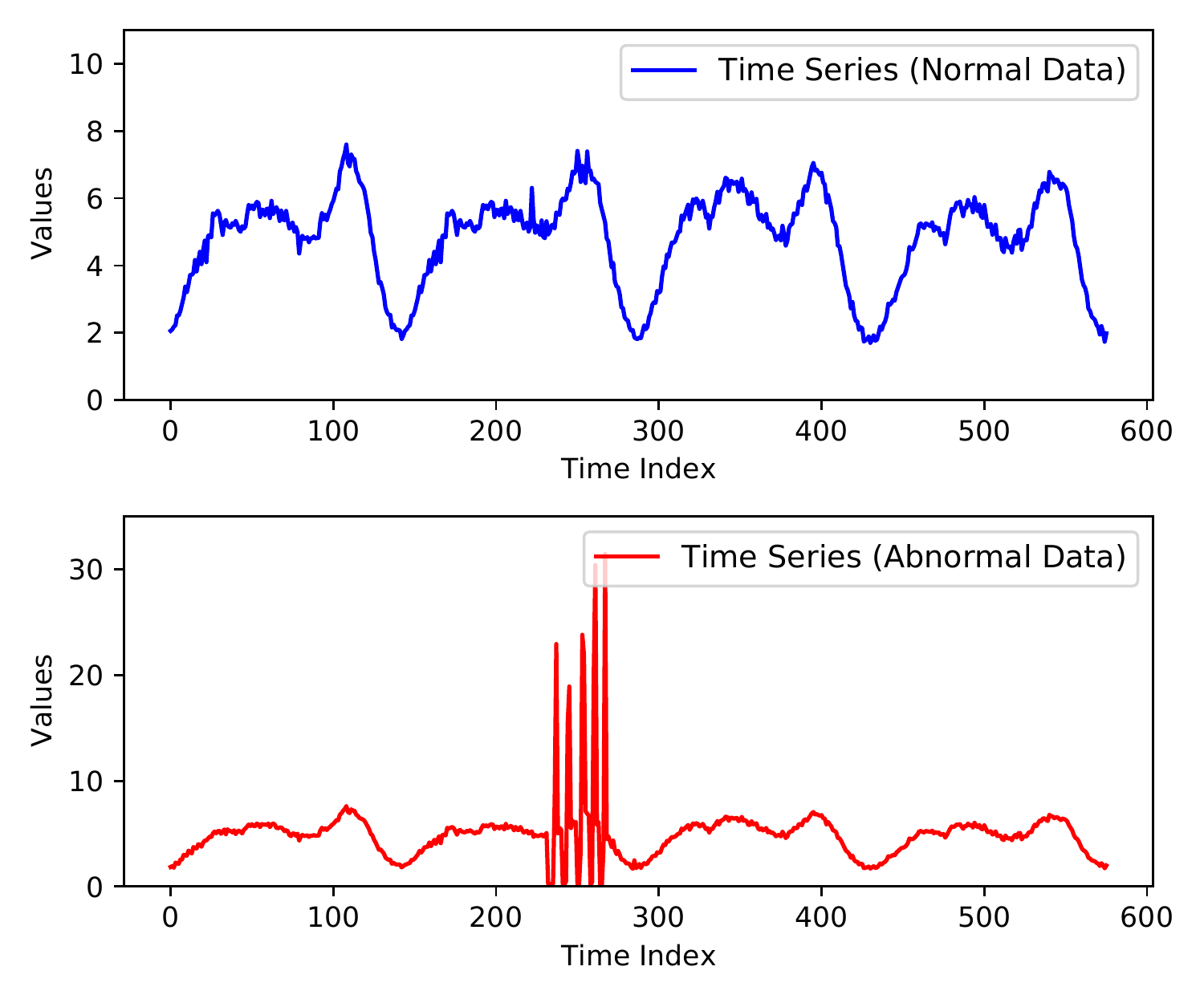}\label{fig:Peri_ACF_oriTS}}
    \subfigure[Original Periodogram and ACF]{\includegraphics[width=0.23\textwidth]{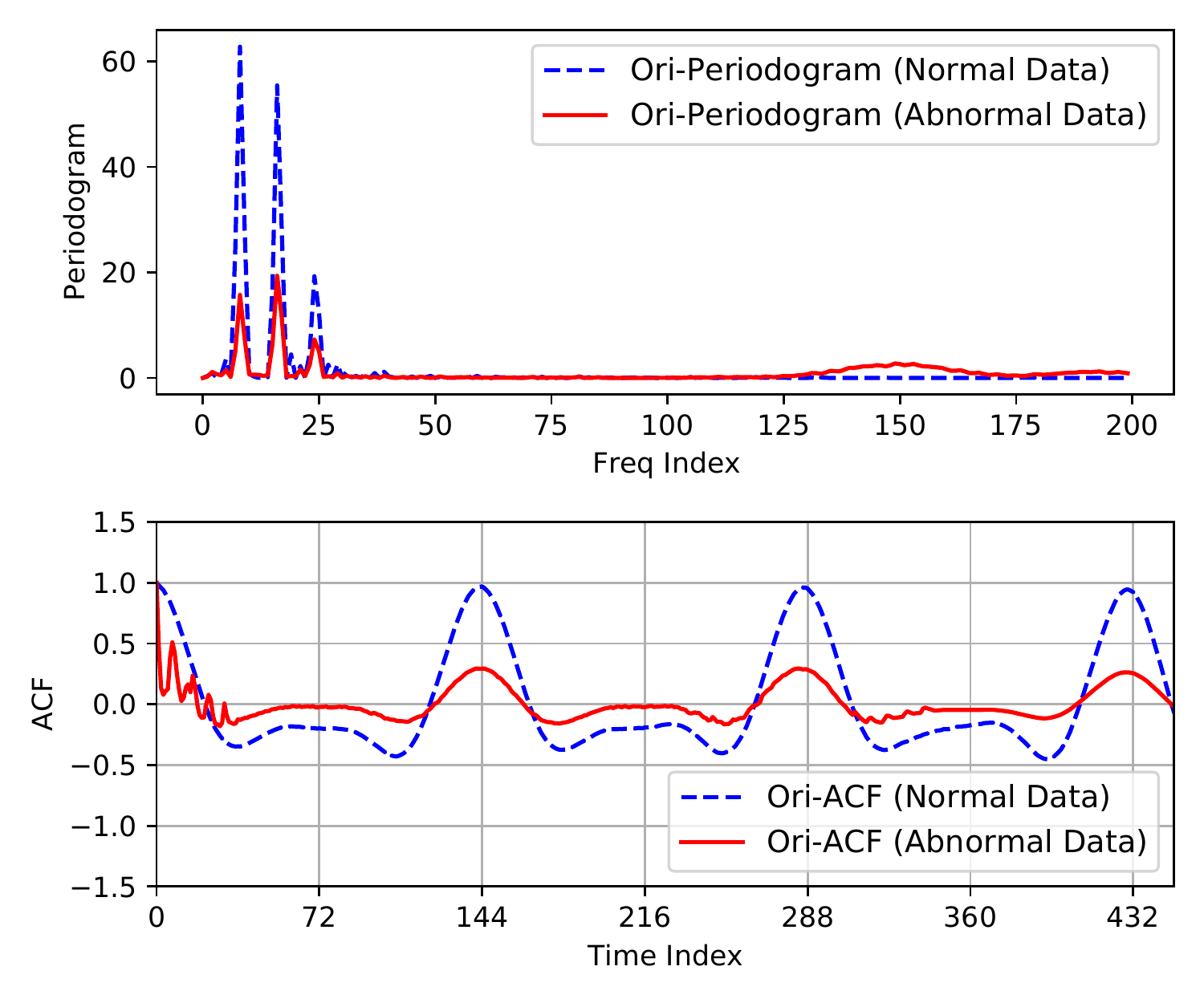}\label{fig:Peri_ACF_Ori}} \\\vspace{-0.2cm}
    \subfigure[LAD-Periodogram and LAD-ACF]{\includegraphics[width=0.23\textwidth]{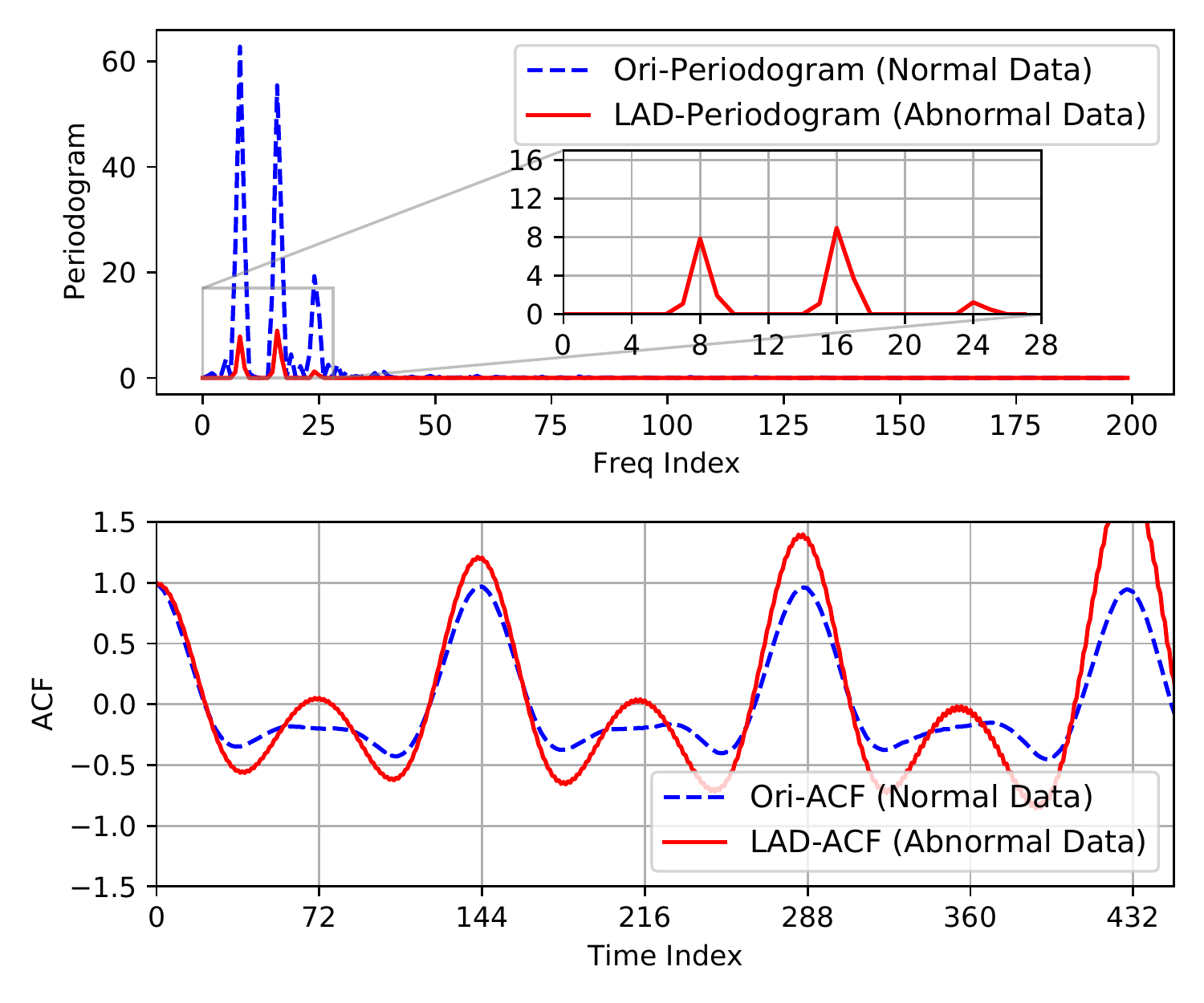}\label{fig:Peri_ACF_LAD}} 
    \subfigure[Huber-Periodogram and Huber-ACF]{\includegraphics[width=0.23\textwidth]{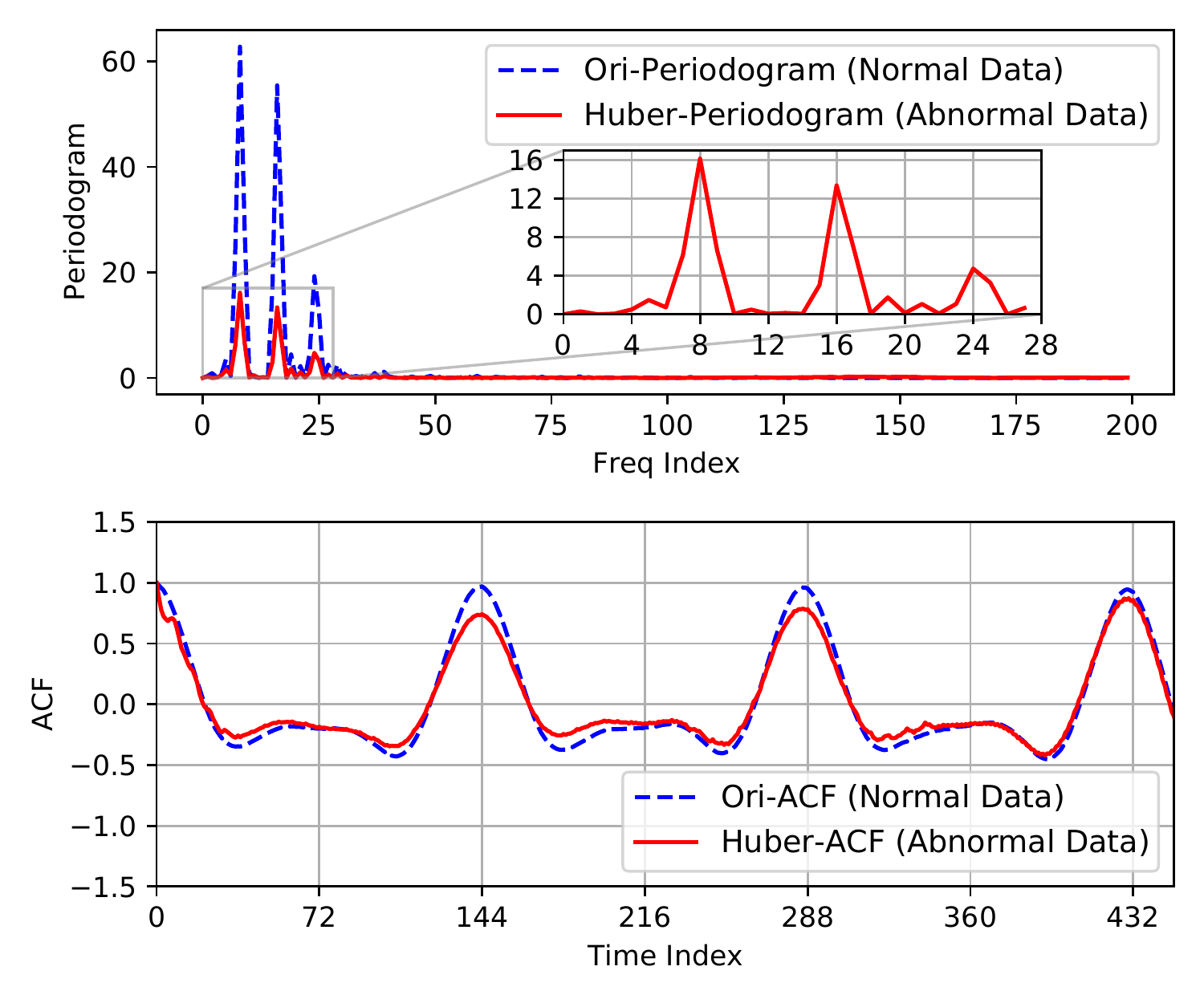}\label{fig:Peri_ACF_Huber}}
    \vspace{-0.4cm}
    \caption{Comparisons of Periodogram and ACF schemes for single-period detection in normal and abnormal cases.}
    \vspace{-0.5cm}
    \label{fig:Peri_ACF_LADHuber}
\end{figure}

\vspace{-0.1cm}
\subsubsection{Effectiveness of Huber-Periodogram and Huber-ACF}
To further understand how the single-periodicity detection works in RobustPeriod algorithm, we show an example in Fig.~\ref{fig:Peri_ACF_LADHuber} based on the real-world Flink Job TPS dataset from Fig.~\ref{fig:realDatasets}. 
% \ref{fig:Peri_ACF_Ori}, Fig.~\ref{fig:Peri_ACF_LAD}, and Fig.~\ref{fig:Peri_ACF_Huber}
The time series of 4-day length (length=576, period=144) in normal (without outliers) and abnormal (with outliers) cases are shown in Fig.~\ref{fig:Peri_ACF_oriTS}.
The outliers severely affect periodogram (e.g., spectral energy at frequency index around 150) and ACF (e.g., the undesirable peaks under 20 in time index) as shown in Fig.~\ref{fig:Peri_ACF_Ori}, which brings difficulties to detect the correct periodicity. 
The use of LAD-periodogram~\cite{Li2008} can somehow obtain better periodogram but the corresponding ACF is still affected as shown in Fig.~\ref{fig:Peri_ACF_LAD}, which would bring the false period length 72. In contrast, our proposed Huber-periodogram and Huber-ACF can obtain similar patterns and the same peak locations as the original periodogram and ACF without outliers as shown in 
Fig.~\ref{fig:Peri_ACF_Huber}, which leads to the correct periodicity detection.

\subsection{Downstream Task with Seasonal Time Series Forecasting}
%(TODO: Liang)

We also evaluate periodicity detection algorithms by considering downstream seasonal time series forecasting task.
% the forecasting performance of seasonal time series using the outputs of different periodicity detection algorithms. 
Specifically, we adopt the state-of-the-art TBATS model (Exponential smoothing state space model with Box-Cox transformation, ARMA errors, Trend and Seasonal components)~\cite{de2011forecasting} for forecasting, 
since it can effectively deal with time series whit complex periodic components. We use the detected periodic length generated from different detection algorithms as the input of TBATS model on the aforementioned 3-periodic (T=12, 24, 168) Yahoo-A4 datasets (total 100 time series). For each time series, the first half length (840 data points) is used for training, while the rest is evaluated for test. We report the average root mean squared error (RMSE) and mean absolute error (MAE) for two different forecasting horizons, which is summarised in Table~\ref{tab:forecasting}.

From Table~\ref{tab:forecasting}, it can be observed  that the Siegel algorithm achieves the worst performance in all settings mainly due to the low recall of the periodicity. The Wavelet-Fisher achieves slightly better performance than the Siegel algorithm. The AUTOPERIOD and our RobustPeriod achieve the best performance. Their forecasting performance is consistent with the periodicity detection performance as summarized in Table~\ref{tab:multi_period_det}. In summary, our RobustPeriod algorithm achieves the best periodicity detection performance, and the best forecasting performance in terms of both RMSE and MAE.

% {\color{red} %for major added contents in revised version
\vspace{-0.2cm}
\subsection{Scalability Studies and Deployment}

\subsubsection{Comparisons of Running Time}
%(TODO: Liang)

To investigate the scalability of different periodicity detection algorithms, we compare the running time on a set of periodic time series data with different lengths. Specifically, we generate synthetic sin-wave time series of length 1000 with three periodic components (with periodic lengths 20, 50, and 100). By applying the sampling technique, we obtain time series with different lengths from 500 to 2000. We randomly generate 1000 time series for each selected length, and then report the average running of different algorithm on a MacBook Pro with Intel i5 2.3GHz CPU and 8GB RAM. 
%first generate synthetic sin-wave 3-periodic time series (T=20,50,100) with length 1000. Then we use 2-times down- and up-sampling to obtain length 500 and 1000 time series, respectively. 
%To obtain the average running time, we randomly generate 1000 time series for each length.
%All the experiments are conducted on a MacBook Pro with a 2.3GHz Intel i5 CPU. 

Table~\ref{tab:diff_len_F1_runningtime} and Table~\ref{tab:diff_len_F1} summarize the average running time and the F1 score of different periodicity detection algorithm as time series length increases from 500 to 2000, respectively. The proposed RobustPeriod algorithm achieves significantly better performance than others at the cost of more running time. From Table~\ref{tab:diff_len_F1_runningtime}, it can be observed that the running time of all algorithm increases as the time series length increases. Also note that all algorithms achieve the periodicity detection task within 1 second. In practice, time series with more length can be down-sampled and tested for periodicity. Compared with other simpler algorithm, our RobustPeriod spends more time, but it is acceptable in real-world applications. Also note that the F1 score decreases significantly as time series length increases for all algorithms except RobustPeriod in Table~\ref{tab:diff_len_F1}. It is due to the increased complexity and interference of noises and outliers. Overall, the proposed RobustPeriod algorithm achieves the best trade-off between detection accuracy and scalability.

\begin{table}[!t] % \begin{table*}[t]
% \small
\centering
\small
%\normalsize
% \tiny, \scriptsize, \footnotesize, \small, \normalsize, \large, \Large, \LARGE, \huge, and \Huge.
% \vspace{-0.2cm}
\caption{Time series forecasting results under different periodicity detection algorithms on Yahoo A-4 datasets (T=12, 24, 168). Forecasting horizon is indicated by h, and the best results are highlighted.}
\vspace{-0.4cm}
\label{tab:forecasting}
% {\color{red}
\begin{tabular}{c|c|c|c|c}
\hline  %          &   
\multirow{2}{*}{Algorithms}  & \multicolumn{2}{|c}{ RMSE } & \multicolumn{2}{|c}{ MAE   } \\ \cline{2-5}
%  &  pr & re & f1     & pr  & re & f1\\ \hline\hline
  &  h=84 & h=168       &   h=84 & h=168    \\ \hline\hline
Siegel  & 430.9  &819.9                 &  268.4         & 440.5         \\  \hline
AUTOPERIOD   & 343.9  & 421.5       & 231.8          &290.9          \\ \hline
Wavelet-Fisher       & 411.8  & 466.1                 &244.9          &274.3         \\ \hline
\textbf{RobustPeriod} & \textbf{334.7}  & \textbf{404.9}    & \textbf{221.7} & \textbf{266.8}  \\ \hline
\end{tabular}
% }
\vspace{-0.4cm}
\end{table}

\begin{table}[t] % \begin{table*}[t]
\centering
\small
%\normalsize
% \tiny, \scriptsize, \footnotesize, \small, \normalsize, \large, \Large, \LARGE, \huge, and \Huge.
% \vspace{-0.2cm}
\caption{Average running time of different periodicity detection algorithms on synthetic data with different lengths.}
\vspace{-0.4cm}
\label{tab:diff_len_F1_runningtime}
% {\color{red}
\begin{tabular}{c|c|c|c}
\hline  %          &   
\multirow{2}{*}{Algorithms}  & \multicolumn{3}{|c}{Time series length}  \\ \cline{2-4}
%  &  pr & re & f1     & pr  & re & f1\\ \hline\hline
  &   Length=500 & Length=1000     &  Length=2000    \\ \hline\hline
Siegel           & 0.003 s &  0.008 s &  0.013 s  \\  \hline
AUTOPERIOD       & 0.014 s      & 0.023 s         &0.046 s  \\ \hline
Wavelet-Fisher    & 0.004 s      &0.006 s          & 0.012 s    \\ \hline
\textbf{RobustPeriod}    & 0.142 s & 0.146 s &  0.300 s  \\ \hline
\end{tabular}
% }
\vspace{-0.2cm}
\end{table}

\begin{table}[t] % \begin{table*}[t]
\centering
% \footnotesize
\small
% \tiny, \scriptsize, \footnotesize, \small, \normalsize, \large, \Large, \LARGE, \huge, and \Huge.
% \vspace{-0.2cm}
\caption{F1 score of different periodicity detection algorithms on synthetic data with different lengths.}
\vspace{-0.4cm}
\label{tab:diff_len_F1}
% {\color{red}
\begin{tabular}{c|c|c|c}
\hline  %          &   
\multirow{2}{*}{Algorithms}  & \multicolumn{3}{|c}{ time series length }  \\ \cline{2-4}
%  &  pr & re & f1     & pr  & re & f1\\ \hline\hline
    & Length=500 & Length=1000     &  Length=2000    \\ \hline\hline
Siegel          & 0.79 &  0.79 &  0.52 \\  \hline
AUTOPERIOD        & 0.79      & 0.25        &0.15 \\ \hline
Wavelet-Fisher   & 0.50      & 0.50          &0.41    \\ \hline
\textbf{RobustPeriod}    & \textbf{0.99} & \textbf{0.99} &  \textbf{0.97}  \\ \hline
\end{tabular}
% }
\vspace{-0.3cm}
\end{table}

\subsubsection{Deployment and Applications}
The proposed RobustPeriod algorithm is implemented and provided as a public online service at one cloud computing company. For the time series within several thousand points length, the running time of the proposed RobustPeriod is usually within 1 second under a regular single-core CPU. Note that in practice the time series periodicity detection is not performed very frequently for most cases. Typically it is performed regularly in a relatively low frequency such as several hours or when a new task is launched. 
The proposed RobustPeriod has been applied widely in different business lines, including AIOps for cloud database and computing, forecasting and anomaly detection for business metrics, and auto-scaling of computing resources.

% \vspace{-0.2cm}
\section{Conclusion}\label{sec:conc}
% \vspace{-0.1cm}

In this paper we propose a new periodicity detection method RobustPeriod by mining periodicities from both time and frequency domains. It utilizes MODWT to isolate the interlaced multiple periodicities successfully. 
To identify the potential periodic pattern at different levels, we apply the robust wavelet variance to select the most promising ones. Furthermore, we adopt Huber-periodogram and the corresponding Huber-ACF to detect periodicity accurately and robustly. The theoretical properties of Huber-periodogram for Fisher's test are also proved. In the future, we plan to apply RobustPeriod in more time series related tasks.
\bibliographystyle{ACM-Reference-Format}
\balance
\bibliography{6_Periodicity_bibfile}

%%

%%% Note: SIGMOD--> No appendix will be allowed.
% \newpage
% \appendix
% \input{8_Supple_rPeriod.tex}

% % \input{7_Appendix.tex}

\end{document}